\documentclass[]{sysu_preprint}

\usepackage[utf8]{inputenc}

\usepackage{url}
\usepackage{amsmath}
\usepackage{amssymb}
\usepackage{amsfonts}
\usepackage{mathtools}
\usepackage{amsthm}
\usepackage{nicefrac}
\usepackage{colortbl}
\usepackage{xcolor}
\usepackage{enumitem}
\usepackage{pifont}
\usepackage{algorithm}
\usepackage{algorithmic}

\usepackage[disable,textsize=tiny]{todonotes}

\definecolor{ourmethod}{RGB}{230,243,255}
\definecolor{datasetcol}{RGB}{240,235,248}
\definecolor{best}{gray}{0.75}
\definecolor{second}{gray}{0.88}

\theoremstyle{plain}

\theoremstyle{definition}

\crefname{assumption}{Assumption}{Assumptions}
\theoremstyle{remark}

\usepackage{type1cm}
\newcommand{\papertablesize}{\fontsize{7.5pt}{9pt}\selectfont}
\newcommand{\method}{\ifmmode\mathrm{G}^{2}\mathrm{MAF}\else G\textsuperscript{2}MAF\fi}
\newcommand{\teaserfigure}{%
  \begin{center}
    \includegraphics[width=0.98\linewidth,height=0.25\textheight,keepaspectratio] {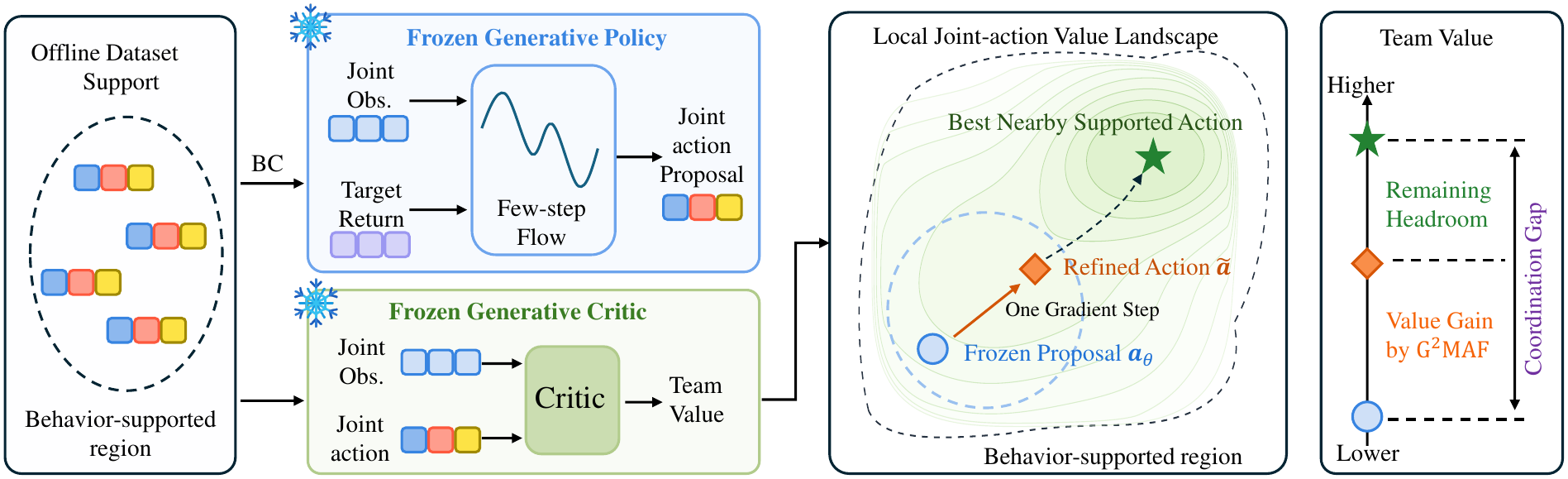}
  \end{center}
  {\captionsetup{font=small}\captionof{figure}{Motivation and \method{} at test time. A frozen generative policy proposes a joint action, and a frozen centralized behavior critic supplies one joint gradient step toward higher predicted team value within the neighborhood supported by the behavior data. The right panel decomposes the samplewise coordination gap into the value gain recovered by \method{} and the remaining local headroom.}\label{fig:gap}}
  \vspace{0.5em}%
}

\title{\method{}: Test-Time Gradient Guidance for Multi-Agent Flow Policies}
\renewcommand{\authorlist}{%
  \authorfont{\sffamily Guowei Zou, Haitao Wang, Guoxin Wang, Zhiquan Chen, Beiwen Zhang, Guojie Wang, Hejun Wu}%
}
\renewcommand{\affiliationlist}{\affiliationfont{\sffamily\seedblue{Sun Yat-sen University}}}

\abstract{Offline multi-agent reinforcement learning (MARL) learns cooperative policies from fixed datasets without further environment interaction and a learned policy is frozen at deployment. Such a frozen policy typically proposes a single joint action and executes it directly at deployment time. However, this one-shot deployment often commits to a suboptimal proposal, even when better nearby alternatives remain consistent with the behavior data. To address this issue, we propose Gradient Guided Multi Agent Flow (\textbf{\method{}}), a refinement framework for optimizing joint policies at test-time. \method{} applies one globally normalized, projected critic gradient to guide and coordinate all agents' corrections while keeping the action both feasible and close to the frozen policy proposal. Across 24 MPE and SMAC settings, its canonical variant improves 20 frozen settings, with mean relative gains of 9.2\% on MPE and 8.9\% on SMAC, with model inference latency increased by about 6\% only.}
\hypersetup{
  pdftitle={G2MAF: Test-Time Gradient Guidance for Multi-Agent Flow Policies},
  pdfauthor={Guowei Zou, Haitao Wang, Guoxin Wang, Zhiquan Chen, Beiwen Zhang, Guojie Wang, Hejun Wu},
  pdfsubject={cs.AI, cs.LG, cs.MA},
  pdfkeywords={multi-agent reinforcement learning, offline reinforcement learning, test-time planning}
}

\definecolor{coverpurple}{HTML}{49308C}
\definecolor{covergray}{HTML}{F1F4F8}
\renewcommand{\maketitle}{%
  \thispagestyle{plain}%
  \begin{tcolorbox}[enhanced,colback=covergray,frame hidden,arc=3mm,
    left=6mm,right=6mm,top=4mm,bottom=4mm,boxsep=0pt,before skip=0pt,after skip=5mm]
    {\raggedright\sffamily\bfseries\color{coverpurple}\fontsize{17}{20}\selectfont
      \titlelist\par}
    \vspace{2mm}
    {\raggedright\renewcommand{\authorfont}{\fontsize{10}{12}\selectfont}
      \bfseries\authorlist\par}
    \vspace{2mm}
    {\raggedright\sffamily\bfseries\color{coverpurple}\fontsize{11}{13}\selectfont Sun Yat-sen University\par}
    \vspace{3mm}
    {\color{coverpurple!30}\hrule height 0.3pt}
    \vspace{1mm}
    \teaserfigure
    \vspace{1mm}
    {\color{coverpurple!30}\hrule height 0.3pt}
    \vspace{2mm}
    {\small\hypersetup{urlcolor=coverpurple}
      {\sffamily\bfseries Project Page:} \url{https://g2maf.github.io/}\par
      {\sffamily\bfseries Code:} \url{https://github.com/g2maf/G2MAF}\par
      {\sffamily\bfseries Models:} \url{https://huggingface.co/g2maf/G2MAF}\par
      {\sffamily\bfseries Datasets:} \url{https://huggingface.co/datasets/Guowei-Zou/CoFlow-datasets}\par}
  \end{tcolorbox}
  {\centering\sffamily\bfseries Abstract\par}
  \vspace{2mm}
  {\setlength{\parindent}{0pt}\abstractlist\par}
  \par\vspace{3mm}
}

\begin{document}
\maketitle

% Offline multi-agent reinforcement learning (MARL) learns cooperative policies from fixed datasets without further environment interaction and a learned policy is frozen at deployment. Such a frozen policy typically proposes a single joint action and executes it directly. However, this one-shot deployment may commit to a suboptimal proposal, even when better nearby alternatives remain consistent with the behavior data. 

\raggedbottom
\section{Introduction}

Offline multi-agent reinforcement learning (MARL) trains joint policies from fixed datasets without further interaction \citep{pan2022omar,formanek2023ogmarl,zhan2025exploiting,liu2025inspo,fu2025ins}.  At deployment, a frozen policy proposes one joint action and executes it directly. Direct execution neither compares nearby alternatives nor revises this first proposal. It retains a local coordination error: actions that seem reasonable individually frequently work poorly together, although a small joint adjustment could raise team return. Generative policies represent joint-action or trajectory distributions \citep{park2025flowql,zhang2025efm,janner2022diffuser,ajay2023dd,zhu2023madiff,yuan2025madits,li2025dof,lee2025macflow,qiao2025score,zou2026coflow}, but one sample does not certify a well-coordinated joint decision. A centralized behavior critic scores a complete joint action at the current joint observation. We call the critic-predicted score difference between the frozen proposal and a higher-scoring nearby joint action the \emph{coordination gap} (Fig.~\ref{fig:gap}).

Bridging this gap requires more than independently adjusting each agent's action. Existing single-agent methods use critic guidance during inference or policy optimization \citep{jang2025q,xu2026vgf,frans2025guidance,li2026adjoint,bagatella2025gcttt}. We examine whether one local, critic-guided joint update improves a frozen cooperative policy without policy retraining, a world model, or candidate rollouts. For few-step generators, the same value signal can be injected before decoding or after the executable action is produced. We treat this location as a design choice.

We propose \textbf{\method{}}, a test-time refinement method with a frozen joint policy and a frozen centralized behavior critic. For each proposal, \method{} differentiates the critic with respect to the complete joint action, normalizes the concatenated gradient once, takes one action-space step, and projects the resulting action onto the feasible set. Because the critic receives every agent's observation and action \citep{lowe2017maddpg,rashid2018qmix}, the resulting correction for each agent depends on its teammates' choices, while the shared normalization gives the team one correction budget. The canonical variant refines the decoded executable action. We additionally study trajectory-space injection for few-step flow policies and masked-logit refinement for discrete actions. Under local smoothness and a sufficiently small step size, the prescribed nonzero continuous-action update strictly increases the critic prediction.

We evaluate \method{} on 24 MPE and SMAC settings.  At the operating points selected for Table~\ref{tab:main}, the canonical variant improves 20 frozen settings, with mean relative gains of 9.2\% on MPE and 8.9\% on SMAC. Paired rollouts measure realized return, critic diagnostics evaluate local ranking, and the injection study compares the three variants. The critic step increases model inference latency by about 6\% on average.

Our contributions are as follows:
\begin{itemize}
\item We formulate the \emph{coordination gap} as the centralized critic's local score difference between a frozen proposal and a higher-scoring nearby joint action.
\item We propose \method{}, which applies one normalized joint-gradient action step followed by feasibility projection, and extend the refinement to trajectory injection and masked-logit discrete control. Under explicit conditions, we prove that the local critic increases or at least equals to that of the frozen policy.
\item We evaluate return gains, injection locations, critic ranking, matched radius controls, and deployment cost across 24 MPE and SMAC settings.
\end{itemize}

\section{Related Work}

\paragraph{Offline MARL.}
Offline MARL learns joint policies from logged data \citep{pan2022omar,formanek2023ogmarl,liu2025inspo,fu2025ins}. Value decomposition methods and centralized training with decentralized execution (CTDE) use joint information during training but execute local policies. QMIX uses monotonic value factorization \citep{rashid2018qmix}, whereas MADDPG uses a centralized critic during training \citep{lowe2017maddpg}. Offline variants add conservatism to combat extrapolation error \citep{pan2022omar}. Recent work also studies interaction aware data synthesis \citep{fu2025ins}, sequential score decomposition of joint behavior \citep{qiao2025score}, and the sensitivity of conclusions in offline MARL to dataset construction \citep{formanek2026data}. \method{} borrows the centralized joint critic architecture but changes its role and information pattern: the critic remains available to a centralized coordinator at test time and refines the complete joint action without training the policy.

\paragraph{Generative policies for multi-agent reinforcement learning.}
Diffusion and flow models have become strong offline policies and planners \citep{park2025flowql,zhang2025efm}. Diffuser and Decision Diffuser address the single-agent case \citep{janner2022diffuser,ajay2023dd}, while multi-agent generative policies include diffusion methods \citep{zhu2023madiff,yuan2025madits,li2025dof,li2026dom2} and flow methods \citep{li2025om2p,lee2025macflow,pang2026vgm2p,zou2026coflow}. Recent single-agent work combines flow policies with energy guidance \citep{alles2025flowq,tiofack2025gfp} and develops one-step or shortcut flow policies \citep{chen2025onestep,espinosa2025shortcut}. We refer to CoFlow as the coordinated generative backbone (CGB) used here. It combines few-step trajectory generation, cross-agent coordination modules, and inverse-dynamics decoding. Its return-conditioned behavior-cloning objective anchors generated actions to offline coordination patterns; \method{} refines the remaining local gap at deployment.

\paragraph{Test-time policy improvement.}
  A growing body of work guides frozen-policy outputs at deployment \citep{jang2025q,xu2026vgf,frans2025guidance}. QAM instead fine-tunes flow-policy parameters with critic action gradients \citep{li2026adjoint}. Other test-time offline RL methods update policy parameters using selected logged transitions \citep{bagatella2025gcttt}; \method{} keeps both networks frozen and edits only the proposed joint action. \method{} studies cooperative value guidance through a centralized joint-action critic. Its trajectory-space and action-space variants share the same frozen policy and critic interface, while few-step multi-agent generation makes the injection point an explicit design choice. Planning with a learned world-action model \citep{mawam2027} is an orthogonal route to test-time deliberation along the temporal axis.

\section{Preliminaries}
\label{sec:prelim}

We use the standard cooperative offline MARL setting. Let $\bm o_{t,i}\in\mathcal O_i$ and $\bm a_{t,i}\in\mathcal A_i$ denote the local observation and action of agent $i\in\{1,\ldots,n\}$ at time $t$. We write the joint observation and action as $\bm o_t=(\bm o_{t,1},\ldots,\bm o_{t,n})\in\mathcal O=\prod_i\mathcal O_i$ and $\bm a_t=(\bm a_{t,1},\ldots,\bm a_{t,n})\in\mathcal A=\prod_i\mathcal A_i$. A fixed dataset $\mathcal D=\{(\bm o_t,\bm a_t,r_t,\bm o_{t+1},c_t)\}$ is collected by an unknown joint behavior policy, where $r_t$ is the shared reward and $c_t$ is the continuation indicator. The frozen generative joint policy samples $\bm a_\theta\sim\pi_\theta(\cdot\mid\bm o_t,g)$ for target return $g$, and the centralized critic scores $Q_\phi(\bm o_t,\bm a_t)$. The reported experiments use centralized test-time coordination execution: one coordinator receives $\bm o_t$, refines all action blocks jointly, and dispatches $\tilde{\bm a}_{t,i}$ to agent $i$; neither model is updated online.  Appendix~B details the Dec-POMDP, coordination gap, execution assumptions, and flow backbone.

\begin{figure*}[t]
\centering
\includegraphics[width=0.98\textwidth]{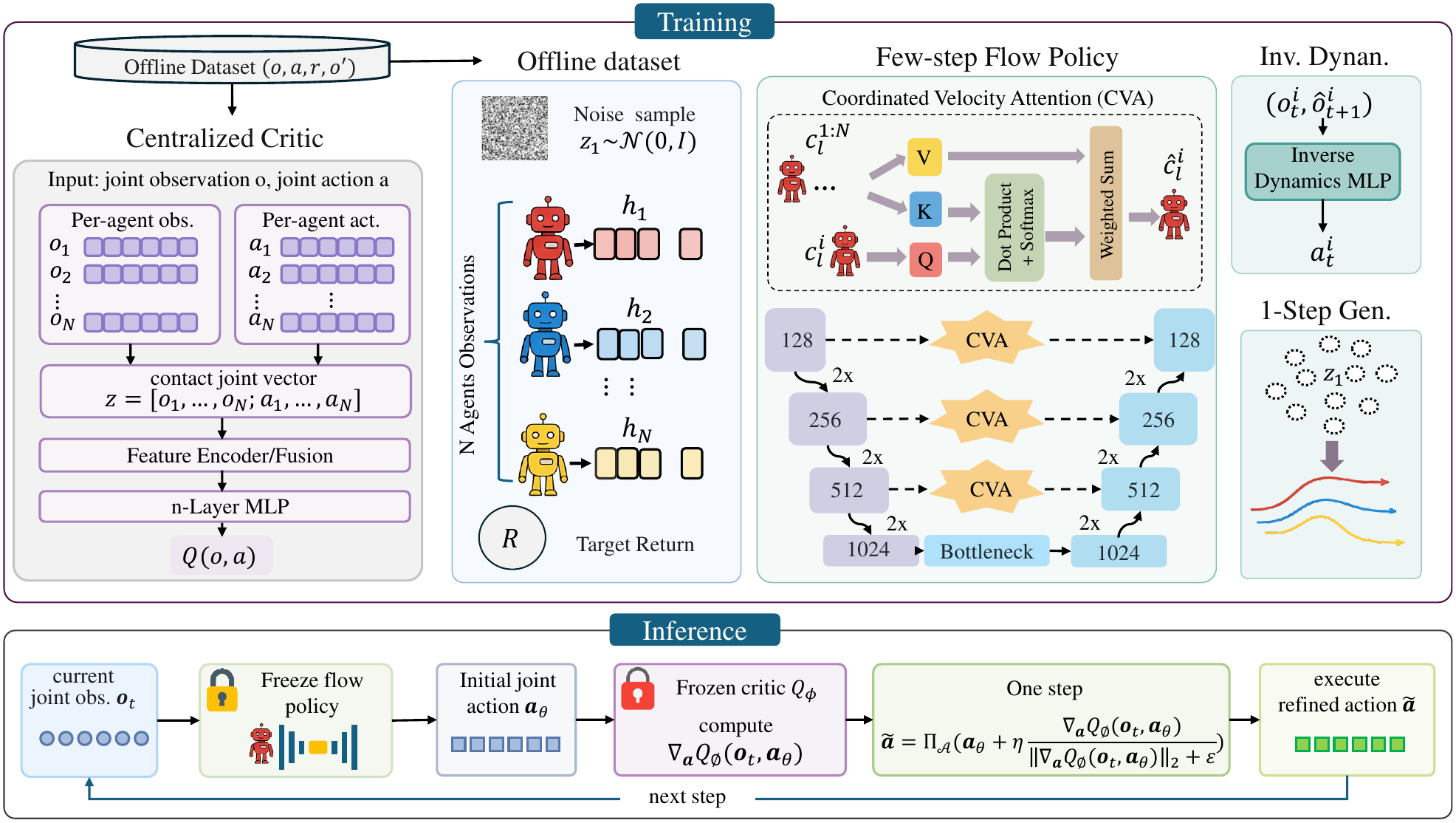}
\caption{Framework of \method{}. Given the joint observation and target return, a frozen generative policy proposes $\bm{a}_\theta$. The centralized critic evaluates the complete joint action and supplies one gradient whose blocks for individual agents depend on all teammates' current actions. \method{} globally normalizes this gradient, projects the refined action $\tilde{\bm{a}}$ onto the feasible action set, and executes it. The method uses neither policy retraining nor rollouts; the SMAC variant applies the same update to logits with illegal actions masked out.}
\label{fig:framework}
\end{figure*}

\section{Method}

\subsection{Framework Overview}
 At each decision, \method{} performs a local edit of a frozen policy proposal. Given the joint observation $\bm o$ and target return $g$, the frozen policy first emits a complete joint action $\bm a_\theta\sim\pi_\theta(\cdot\mid\bm o,g)$. A centralized critic then differentiates its predicted team value with respect to every component of $\bm a_\theta$. \method{} applies one bounded update to the full joint vector and executes the resulting feasible action $\tilde{\bm a}$. Thus the policy supplies a behavior-like starting point, while the critic only chooses a local correction; neither model is updated at deployment.

\subsection{Coordinated Flow Prior}
The frozen proposal policy is a coordinated flow model. It transports a normalized joint observation trajectory from noise $\bm z_1\sim\mathcal N(0,I)$ toward the data trajectory $\bm x_0$ along $\bm z_\alpha=(1-\alpha)\bm x_0+\alpha\bm z_1$. Let $\bm c=(\mathsf N_o(\bm o),g)$ denote the observation and return condition. Its velocity network is trained by the flow matching objective:
\begin{equation}
\mathcal{L}_{\mathrm{flow}}(\vartheta)
=
\mathbb{E}_{\alpha,\bm{x}_0,\bm{z}_1}
\Big[
\big\|u_\vartheta(\bm{z}_\alpha,0,\alpha;\bm c)-(\bm{z}_1-\bm{x}_0)\big\|_2^2
\Big].
\label{eq:coflow_loss}
\end{equation}
Cross agent attention makes this velocity a joint function rather than a collection of independent agent flows. A shared inverse dynamics head decodes the final trajectory into $\bm a_\theta$. The backbone is trained before refinement and remains frozen;  Appendix~B gives its sampler and decoder.

\subsection{Centralized Critic as a Coordination Signal}
$Q_\phi(\bm o,\bm a)$ estimates the team return after taking joint action $\bm a$ at $\bm o$ and then following the logged behavior. Because it receives the complete joint observation--action pair, $\nabla_{\bm a_i}Q_\phi(\bm o,\bm a)$ depends on every teammate's action and therefore specifies a coordinated correction. We train this critic once by behavior fitted critic evaluation (FQE) with one step temporal difference targets and freeze it.  Appendix~B gives the temporal difference (TD) objective; the experiments test whether the critic ranks nearby executable actions consistently with simulator outcomes.

\subsection{Joint Value Guidance at Test Time}
For fixed $(\bm o,g)$, the ideal policy target raises predicted team value while remaining close to the frozen joint prior:
\begin{equation}
\begin{aligned}
\pi_Q
&\in\arg\max_{\pi\ll\pi_\theta}\ \mathcal J_{\bm o,g}(\pi),\\
\mathcal J_{\bm o,g}(\pi)
&=\mathbb E_{\bm a\sim\pi(\cdot\mid\bm o,g)}
 [Q_\phi(\bm o,\bm a)]\\
&\quad-\beta D_{\rm KL}\!\left(
\pi(\cdot\mid\bm o,g)\,\|\,\pi_\theta(\cdot\mid\bm o,g)
\right).
\end{aligned}
\label{eq:marl_kl_objective}
\end{equation}
Here $D_{\rm KL}$ is the Kullback--Leibler (KL) divergence.
Under the regularity conditions in  Appendix~B, its solution is the value tilted joint distribution:
\begin{equation}
\begin{aligned}
Z_\beta(\bm o,g)
&=\int_{\mathcal A}\pi_\theta(\bm a\mid\bm o,g)
\exp\!\left(\frac{Q_\phi(\bm o,\bm a)}{\beta}\right)d\bm a,\\
\pi_Q(\bm a\mid\bm o,g)
&=\frac{\pi_\theta(\bm a\mid\bm o,g)
\exp(Q_\phi(\bm o,\bm a)/\beta)}{Z_\beta(\bm o,g)}.
\end{aligned}
\label{eq:multi_qgf_policy}
\end{equation}
For continuous densities, this tilt adds the centralized value gradient to the prior score:
\begin{equation}
\begin{aligned}
\nabla_{\bm a}\log\pi_Q(\bm a\mid\bm o,g)
&=\nabla_{\bm a}\log\pi_\theta(\bm a\mid\bm o,g)\\
&\quad+\beta^{-1}\nabla_{\bm a}Q_\phi(\bm o,\bm a).
\end{aligned}
\label{eq:multi_qgf_score}
\end{equation}
The CGB policy is an implicit pushforward. \method{} therefore uses the value gradient term directly to refine one sampled decoded joint action, rather than evaluating this score or fitting $\pi_Q$.

\subsection{\method{}: Refinement after Generation}
For continuous control, \method{} directly applies one globally normalized, projected update from the centralized critic to the executable joint action:
\begin{equation}
\tilde{\bm a}
=\Pi_{\mathcal A}\!\left(
\bm a_\theta+
\eta\frac{\nabla_{\bm a}Q_\phi(\bm o,\bm a_\theta)}
{\|\nabla_{\bm a}Q_\phi(\bm o,\bm a_\theta)\|_2+\varepsilon}
\right).
\label{eq:ccr}
\end{equation}
Here $\mathcal A=\prod_i\mathcal A_i$ is the feasible box for the joint action, $\Pi_{\mathcal A}$ is Euclidean projection, and $\varepsilon=10^{-6}$ prevents numerical division by zero. The derivative is evaluated on the complete joint action. Each agent's correction therefore depends on its teammates' current actions, and shared normalization gives the team one correction budget.  Appendix~B defines the blocks for individual agents and gives the trust region derivation, displacement bound, and local critic ascent condition. Each decision uses one critic backward pass and no actor update, candidate search, or model rollout.

\subsection{Where to Inject the Value Signal}
\method{} supports value-signal injection before or after the trajectory decoder. For CGB, the variable before decoding is a normalized joint observation trajectory. A trajectory variant first takes one reverse Euler step and decodes its observation--action pair,
\begin{equation}
\begin{aligned}
\bar{\bm z}^{k-1}
&=\bm z^k-\Delta\alpha_k\,
u_\vartheta(\bm z^k,0,\alpha_k;\mathsf N_o(\bm o),g),\\
\bm x_{\rm c}^{k-1}&=\mathcal C(\bar{\bm z}^{k-1};\mathsf N_o(\bm o)),\\
\bm y^{k-1}&=\left(P_{H_{\rm hist}}\bm x_{\rm c}^{k-1},
\mathcal I_\psi(P_{H_{\rm hist}}\bm x_{\rm c}^{k-1},
P_{H_{\rm hist}+1}\bm x_{\rm c}^{k-1})\right).
\end{aligned}
\label{eq:coflow_intermediate_action}
\end{equation}
Its critic in normalized coordinates supplies the chain rule direction through $J_{\rm dec}^{k-1}=\partial\bm y^{k-1}/\partial\bar{\bm z}^{k-1}$:
\begin{equation}
\begin{aligned}
\bm G_{\rm traj}^{k-1}
&=\nabla_{\bar{\bm z}^{k-1}}Q_\nu^{\rm norm}(\bm y^{k-1})\\
&=\left(J_{\rm dec}^{k-1}\right)^\top
\nabla_{\bm y^{k-1}}Q_\nu^{\rm norm}(\bm y^{k-1}).
\end{aligned}
\label{eq:coflow_chain_guidance}
\end{equation}
 All applies this update after every reverse Euler step; Final applies it only at $k=1$. Post completes the flow, decodes $\bm a_\theta$, and applies Eq.~\ref{eq:ccr} directly in action space. Thus all three are \method{} variants distinguished by the injection point: trajectory variants propagate the critic gradient through the decoder Jacobian, while Post uses the action gradient after decoding. A nonlinear decoder generally makes the two edits nonequivalent.  Appendix~B gives the complete trajectory update.
\subsection{Discrete Joint  Actions}
For SMAC, agent $i$ emits logits $\bm\ell_i\in\mathbb R^{A_i}$ and receives a mask of legal actions $\bm m_i\in\{0,1\}^{A_i}$. We apply softmax separately over each agent's legal actions, differentiate the critic through the resulting relaxation, and mask again before selecting the discrete action:
\begin{equation}
\begin{aligned}
\bm p_i&=\operatorname{softmax}(\bm\ell_i+\log\bm m_i),
\qquad \bm p=(\bm p_1,\ldots,\bm p_n),\\
\bm d_\ell&=\nabla_{\bm\ell}Q_\phi(\bm o,\bm p),
\qquad D_\varepsilon^\ell=\|\bm d_\ell\|_2+\varepsilon,\\
\tilde{\bm\ell}_i&=\bm\ell_i+
\eta\frac{\bm d_{\ell,i}}{D_\varepsilon^\ell},
\qquad
\tilde a_i=\arg\max_{1\le j\le A_i}
\{\tilde\ell_{ij}+\log m_{ij}\}.
\end{aligned}
\label{eq:discrete}
\end{equation}
We use $\log0=-\infty$. Masking again before $\arg\max$ prevents an illegal action from being executed. The concatenated logits share one normalization, with $\|\tilde{\bm\ell}-\bm\ell\|_2\leq\eta$. A discrete action changes only when competing logits cross; its useful step sizes in logit space are therefore larger than the radii for continuous actions.

\begin{table*}[!t]
\centering
\papertablesize
\par\vspace{0.5em}\noindent \textbf{(a) MPE, OMAR-normalized score}\par\vspace{0.25em}
{\papertablesize\setlength{\tabcolsep}{0.6mm}
\begin{tabular*}{\textwidth}{@{\extracolsep{\fill}}llcccccccccc}
\toprule
Task & Qual. & Data & BC & ICQ & TD3+BC & CQL & OMAR & Diff & MA-SfBC & DOM2 & \method{} \\
\midrule
 \multirow{4}{*}{Spread} & Expert & 96.2 & 35.0$\pm$2.6 & 104.0$\pm$3.4 & 108.3$\pm$3.9 & 98.2$\pm$5.2 & \underline{114.9$\pm$2.6} & 95.0$\pm$5.3 & 87.5$\pm$7.3 & 88.7$\pm$6.3 & \textbf{115.2$\pm$0.9} \\
  & MedRep & 9.3 & 10.0$\pm$3.8 & 13.6$\pm$5.7 & 15.4$\pm$5.6 & 31.4$\pm$7.2 & 37.9$\pm$6.1 & 30.3$\pm$2.5 & 8.2$\pm$4.6 & \textbf{63.1$\pm$9.5} & \underline{47.3$\pm$2.2} \\
  & Medium & 27.0 & 31.6$\pm$4.8 & 29.3$\pm$5.5 & 39.4$\pm$3.6 & 34.1$\pm$7.2 & 47.9$\pm$18.9 & \underline{64.9$\pm$7.7} & 51.6$\pm$14.2 & \textbf{78.6$\pm$8.1} & 59.7$\pm$1.7 \\
  & Random & 0.1 & -0.5$\pm$3.2 & 6.3$\pm$3.5 & 9.8$\pm$4.9 & 24.0$\pm$9.8 & 34.4$\pm$5.3 & 6.9$\pm$3.1 & 5.1$\pm$3.9 & \underline{37.4$\pm$11.3} & \textbf{58.0$\pm$9.0} \\
\midrule
 \multirow{4}{*}{Tag} & Expert & 89.5 & 40.0$\pm$9.6 & 113.0$\pm$14.4 & 115.2$\pm$12.8 & 119.3$\pm$14.0 & \underline{123.9$\pm$10.5} & 103.0$\pm$12.0 & 77.4$\pm$13.9 & 98.2$\pm$14.4 & \textbf{139.7$\pm$1.1} \\
  & MedRep & 8.6 & 0.9$\pm$1.4 & 34.5$\pm$27.8 & 28.7$\pm$20.9 & 41.7$\pm$15.3 & 47.1$\pm$15.3 & 53.9$\pm$11.4 & 12.7$\pm$7.3 & \underline{68.2$\pm$16.7} & \textbf{73.5$\pm$7.5} \\
  & Medium & 43.8 & 22.5$\pm$1.8 & 63.3$\pm$20.0 & 65.1$\pm$29.5 & 61.7$\pm$23.1 & 66.7$\pm$23.2 & 72.7$\pm$9.4 & 47.1$\pm$17.9 & \underline{82.6$\pm$18.2} & \textbf{112.7$\pm$1.1} \\
  & Random & 0.0 & 1.2$\pm$0.5 & 2.2$\pm$1.3 & 6.0$\pm$2.1 & 11.1$\pm$2.8 & 11.1$\pm$2.8 & 4.6$\pm$2.6 & 11.6$\pm$5.1 & \underline{29.6$\pm$8.1} & \textbf{46.1$\pm$2.9} \\
\midrule
 \multirow{4}{*}{World} & Expert & 93.8 & 33.0$\pm$9.9 & 109.5$\pm$22.8 & 110.3$\pm$21.3 & \underline{119.8$\pm$28.1} & 110.4$\pm$25.7 & 109.3$\pm$15.4 & 97.3$\pm$19.1 & 99.5$\pm$17.1 & \textbf{156.4$\pm$5.2} \\
  & MedRep & 11.2 & 2.3$\pm$1.5 & 12.0$\pm$9.1 & 17.4$\pm$8.1 & 19.3$\pm$18.3 & 42.9$\pm$19.5 & 19.8$\pm$6.2 & 9.1$\pm$5.9 & \textbf{65.9$\pm$10.6} & \underline{45.8$\pm$2.0} \\
  & Medium & 54.5 & 25.3$\pm$2.0 & 71.9$\pm$20.0 & 73.4$\pm$9.3 & 58.6$\pm$11.2 & 74.6$\pm$11.5 & \underline{84.7$\pm$12.3} & 54.2$\pm$22.7 & 84.5$\pm$23.4 & \textbf{146.8$\pm$11.4} \\
  & Random & 0.0 & -2.4$\pm$0.5 & 1.0$\pm$3.2 & 2.8$\pm$5.5 & 0.6$\pm$2.0 & \underline{5.9$\pm$5.2} & \textbf{6.1$\pm$2.4} & 3.1$\pm$1.3 & 4.1$\pm$1.1 & 5.0$\pm$0.3 \\
\bottomrule
\end{tabular*}}
\par\vspace{0.5em}\noindent \textbf{(b) SMAC, shaped episode return}\par\vspace{0.25em}
{
\papertablesize
\setlength{\tabcolsep}{0.5mm}
\begin{tabular*}{\textwidth}{@{\extracolsep{\fill}}llccccccccccc}
\toprule
Task & Qual. & Data & BC & ICQ & CQL & MA-DT & Diff & DoF & Flow BC & MAC-Flow & VGM$^2$P & \method{} \\
\midrule
\multirow{3}{*}{3m}
 & Good & 16.5 & 16.0$\pm$1.0 & 18.8$\pm$0.6 & 19.0$\pm$0.3 & 19.6$\pm$0.7 & 19.3$\pm$0.5 & 19.8$\pm$0.2 & \textbf{20.0$\pm$0.0} & 19.8$\pm$0.2 & 19.5$\pm$0.7 & \underline{20.0$\pm$0.1} \\
 & Medium & 10.0 & 8.2$\pm$0.8 & 18.1$\pm$0.7 & \underline{18.9$\pm$0.7} & 17.2$\pm$0.7 & 16.4$\pm$2.6 & 18.6$\pm$1.2 & 14.7$\pm$1.5 & 18.0$\pm$3.2 & 16.9$\pm$1.1 & \textbf{21.4$\pm$1.8} \\
 & Poor & 4.7 & 4.4$\pm$0.1 & 14.4$\pm$1.2 & 5.8$\pm$0.4 & 8.9$\pm$0.3 & 10.3$\pm$6.1 & 10.9$\pm$1.1 & 4.5$\pm$0.1 & 10.6$\pm$2.2 & \textbf{14.9$\pm$1.5} & \underline{14.8$\pm$0.9} \\
\midrule
\multirow{3}{*}{2s3z}
 & Good & 18.3 & 18.2$\pm$0.4 & 19.6$\pm$0.3 & 19.1$\pm$0.8 & 19.4$\pm$0.1 & 15.9$\pm$1.2 & 18.5$\pm$0.8 & 19.5$\pm$0.1 & 19.5$\pm$0.5 & \underline{19.9$\pm$0.1} & \textbf{21.2$\pm$0.1} \\
 & Medium & 12.6 & 14.3$\pm$0.7 & 17.2$\pm$0.8 & 14.3$\pm$2.0 & 17.4$\pm$0.3 & 15.6$\pm$0.3 & \underline{18.1$\pm$0.9} & 15.1$\pm$2.0 & 17.6$\pm$0.6 & 16.5$\pm$0.6 & \textbf{19.1$\pm$0.4} \\
 & Poor & 6.9 & 6.7$\pm$0.3 & \underline{12.1$\pm$0.4} & 10.1$\pm$0.7 & 9.9$\pm$0.2 & 8.5$\pm$1.3 & 10.0$\pm$1.1 & 6.9$\pm$0.8 & 8.5$\pm$0.6 & 7.9$\pm$0.7 & \textbf{12.2$\pm$0.3} \\
\midrule
\multirow{3}{*}{5m\_vs\_6m}
 & Good & 16.6 & 16.6$\pm$0.6 & 16.3$\pm$0.9 & 13.8$\pm$3.1 & 18.0$\pm$1.0 & 16.5$\pm$2.8 & 17.7$\pm$1.1 & 14.7$\pm$2.1 & \underline{18.6$\pm$3.5} & 17.6$\pm$1.3 & \textbf{19.4$\pm$0.6} \\
 & Medium & 12.6 & 14.2$\pm$0.5 & 17.2$\pm$0.4 & 16.8$\pm$3.1 & \underline{17.5$\pm$0.4} & 15.2$\pm$2.6 & 16.2$\pm$0.9 & 12.8$\pm$0.8 & 15.6$\pm$1.3 & 17.0$\pm$0.9 & \textbf{20.6$\pm$0.7} \\
 & Poor & 7.5 & 7.5$\pm$0.2 & 9.4$\pm$0.4 & 10.4$\pm$1.0 & 8.9$\pm$0.3 & 8.9$\pm$1.3 & \underline{10.8$\pm$0.3} & 7.7$\pm$0.8 & 9.8$\pm$2.1 & 10.7$\pm$1.1 & \textbf{12.4$\pm$0.6} \\
\midrule
\multirow{3}{*}{8m}
 & Good & 16.9 & 16.7$\pm$0.4 & 19.6$\pm$0.3 & 13.1$\pm$6.1 & 19.2$\pm$0.1 & 18.9$\pm$1.1 & 19.6$\pm$0.3 & 19.5$\pm$0.2 & \underline{19.7$\pm$0.3} & 19.7$\pm$0.4 & \textbf{21.1$\pm$0.3} \\
 & Medium & 10.1 & 10.7$\pm$0.5 & 18.6$\pm$0.5 & 16.3$\pm$3.1 & 18.0$\pm$0.5 & 16.8$\pm$1.6 & 18.6$\pm$0.8 & 18.2$\pm$0.8 & \underline{19.4$\pm$0.6} & 18.2$\pm$1.6 & \textbf{19.7$\pm$0.4} \\
 & Poor & 5.3 & 5.3$\pm$0.1 & 10.8$\pm$0.8 & 4.6$\pm$2.4 & 5.1$\pm$0.1 & 9.8$\pm$0.9 & \textbf{12.0$\pm$1.2} & 4.9$\pm$0.1 & \underline{11.5$\pm$0.8} & 4.9$\pm$0.1 & 9.5$\pm$0.1 \\
\bottomrule
\end{tabular*}
}
\caption{\textbf{Main results} on (a)~MPE and (b)~SMAC. ``Data'' is the offline data mean, ``Diff'' abbreviates MADiff, and \method{} denotes the refined frozen backbone at its selected operating point. Bold and underline mark the best and second best method in each row, excluding Data.  Appendix~A.1 gives the reporting conventions.}
\label{tab:main}
\end{table*}

\section{Experiments}

The experiments study three questions. (Q1) Does \method{} improve frozen offline cooperative policies, and when do realized local headroom and critic reliability translate into gain? (Q2) How does the injection location behave for coordinated flow with few steps? (Q3) Does \method{} transfer to discrete joint actions, and how does its deployment cost compare with world model planning?

\subsection{Setup}
\paragraph{Policies and benchmark.} We evaluate frozen CGB checkpoints. MPE comprises \texttt{simple-spread} cooperative navigation and \texttt{simple-tag}/\texttt{simple-world} predator prey coordination. SMAC comprises \texttt{3m}, \texttt{2s3z}, \texttt{5m\_vs\_6m}, and \texttt{8m} discrete micromanagement. Both benchmarks span dataset quality from \texttt{Poor} to \texttt{Expert}. For MPE predator prey, refinement applies to the three learned predator agents; the prey remains scripted. All $24$ task and quality settings are reported, with representative benchmark trajectories shown in Figure~\ref{fig:dataset-keyframes} (in Appendix).

\paragraph{Baselines.} BC, ICQ, TD3, and CQL denote behavior cloning, implicit constraint Q-learning, Twin Delayed Deep Deterministic Policy Gradient, and conservative Q-learning. The shared baseline panel follows CoFlow's matched comparison~\citep{zou2026coflow} and retains the original-method citations~\citep{pomerleau1991efficient,yang2021believe,fujimoto2021minimalist,kumar2020conservative,pan2022omar,meng2021offline,zhu2023madiff,li2025dof}. These citations identify algorithm origins rather than the source of every multi-agent row. MA-SfBC adapts SfBC~\citep{chen2023sfbc}, with values reported by DOM2~\citep{li2026dom2}. The other newer columns are MAC-Flow~\citep{lee2025macflow} and VGM$^2$P~\citep{pang2026vgm2p}. Each published column retains its reporting source's benchmark split, score scale, and uncertainty convention.

\paragraph{Protocol.} Table~\ref{tab:main} reports the sweep frontier for each setting, and Figure~\ref{fig:cont-step-sweeps} shows sensitivity to the step size. MPE uses OMAR normalized score and SMAC uses shaped episodic return; SMAC success is reported separately. We define $\Delta_R=R_{\mathrm{\method{}}}-R_{\mathrm{Frozen}}$ and, when $R_{\mathrm{Frozen}}\ne0$, $\delta_{\mathrm{rel}}=\Delta_R/|R_{\mathrm{Frozen}}|$ within the displayed benchmark metric. Standard \method{} evaluations aggregate five independent random seeds;  Appendix~A gives the grids, episode budgets, and control procedures.

\subsection{Q1: Gains Track Local Headroom}

\begin{figure*}[!t]
\centering
\includegraphics[width=\textwidth]{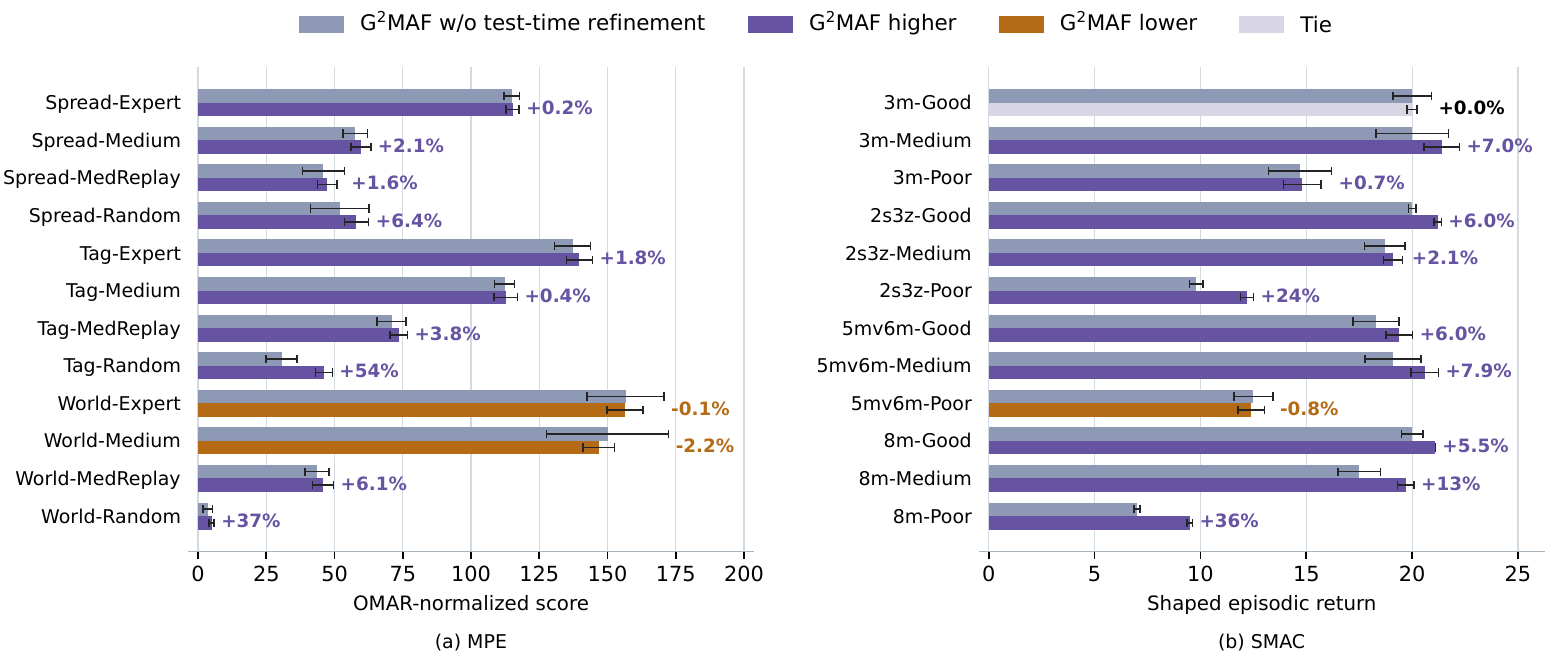}
\caption{\textbf{\method{} gain for each setting} over \method{} without refinement at test time, evaluated at Table~\ref{tab:main} operating points. Colors denote higher, lower, or tied relative gain $\delta_{\mathrm{rel}}$. (a)~MPE OMAR normalized score; (b)~SMAC shaped episodic return. Labels give $\delta_{\mathrm{rel}}$; error bars show the reported uncertainty for \method{} without refinement and $\pm1$ standard error for \method{}.}
\label{fig:main-gains}
\par\vspace{6pt}
\includegraphics[width=\textwidth]{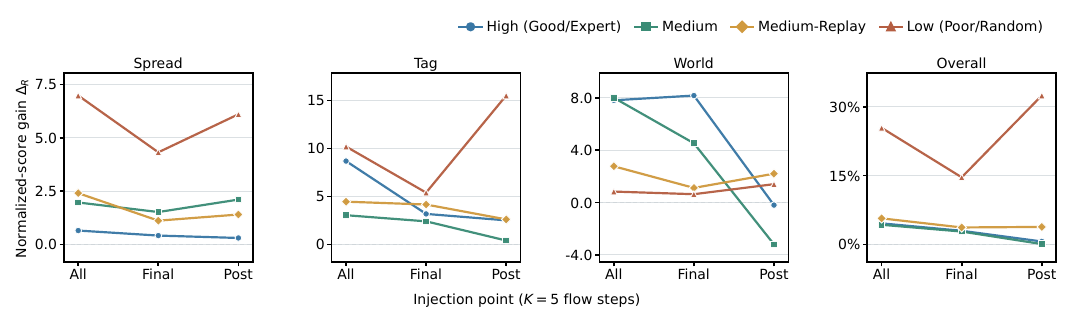}
\caption{\textbf{\method{} gains from different injection locations on MPE.} All $12$ continuous action settings. ``All'' and ``Final'' are independently evaluated trajectory variants with $K{=}5$; ``Post'' is the standard decoded-action gain from Table~\ref{tab:main}. The first three panels show OMAR normalized score gain; the fourth shows mean relative gain by dataset quality.}
\label{fig:q2-q3-dashboard}
\end{figure*}

Table~\ref{tab:main} compares \method{} with published baselines, and Figure~\ref{fig:main-gains} shows its paired gain over the frozen backbone. \method{} improves $20/24$ settings. Gains are largest on several lower-quality MPE and SMAC datasets, whereas nearly saturated settings such as \texttt{simple-spread Expert} and \texttt{3m-Good} change little. Figure~\ref{fig:gain-scatter} (in Appendix) groups these gains by dataset quality.

Three supplementary checks bound this result. Leave-one-task-out (LOTO) step transfer is positive on $18/24$ settings; continuous updates remain local relative to empirical nearest-neighbor action distances; and the critic ranks realized simulator returns above chance on $23/24$ settings. The largest gains span multiple observed headroom--reliability cells rather than concentrating in the high/high cell.  Appendix~A.1 details the locality and critic protocols, and Table~\ref{tab:transfer-controls}(a) (in Appendix) reports step transfer.

\paragraph{ ggregate view and headroom.}
Table~\ref{tab:smac-win} (in Appendix) shows that \method{} raises success rate on several Medium/Poor maps by up to $+21.1$ percentage points while maintaining the strong performance of Good maps. Figure~\ref{fig:gain-quality} (in Appendix) summarizes Figure~\ref{fig:main-gains} gains by dataset quality within each benchmark.

\FloatBarrier

\subsection{Q2: The Injection Point Matters}
Figure~\ref{fig:q2-q3-dashboard} reports all $12$ continuous MPE settings. The trajectory variant that injects at every step is largest on $8$ settings and positive on all $12$; the final-step trajectory variant is largest on $1$ setting and positive on all $12$; the standard variant that refines actions after generation is largest on $3$ settings and positive on $10$. All/Final and Post use different evaluation batches. These counts therefore summarize separate variant evaluations. We deploy Post because it refines the executable action and reuses the behavior critic instead of training a critic in normalized coordinates.

In the paired full grid rerun in Table~\ref{tab:mechanism}(a) (in Appendix), the \method{} gain is positive on $22/24$ settings. Its mean gain exceeds the best of 16 candidates on $18/24$ settings and the one agent update on $19/24$. On the $12$ continuous MPE settings, \method{} exceeds the factorized critic update on $7/12$; this comparison is mixed across settings. Table~\ref{tab:mechanism}(b) (in Appendix) records positive change in critic value on all $24$ settings. For continuous actions, the realized displacement tracks the prescribed radius.

\paragraph{Perturbation controls.}
Table~\ref{tab:transfer-controls}(b) (in Appendix) compares the canonical Table~\ref{tab:main} \method{} gain with independently measured matched nominal-radius Random and Shuffled effects. The separate evaluation batches make these descriptive effect comparisons.

\subsection{Q3: Discrete  Actions and Deployment Cost}

\begin{figure*}[!t]
\centering
\includegraphics[width=0.92\textwidth]{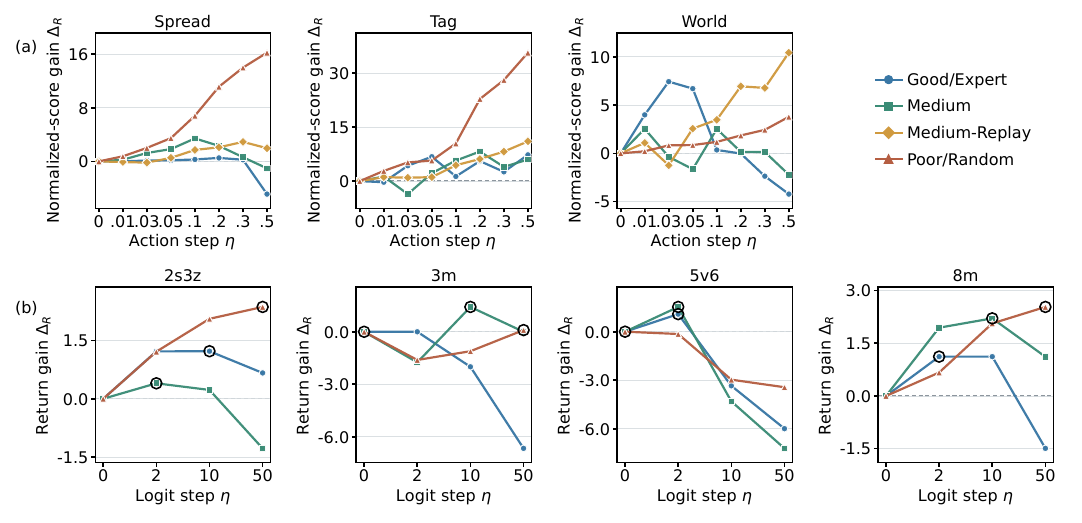}
\caption{\textbf{Sensitivity to the step size across action spaces.} Panel (a) shows the unified grid for continuous actions in OMAR normalized MPE score; panel (b) shows the SMAC grid for masked logits in shaped episodic return. Each continuous marker uses one fixed $K{=}5$ protocol. Curves report gain over their matched Base evaluation, axes are specific to each task, and rings mark the best discrete step.  Appendix~A.3 gives the full protocol.}
\label{fig:cont-step-sweeps}
\end{figure*}

In discrete control, \method{} differentiates the masked logits before action selection. The sensitivity rerun over 12 settings in Figure~\ref{fig:cont-step-sweeps}(b) has positive sweep frontier gain on $10/11$ settings that are not saturated; the useful $\eta$ varies by map and dataset quality because an action changes only when masked logits cross.

Table~\ref{tab:latency} (in Appendix) reports deployment cost. Over all $24$ settings, the critic gradient takes $3.85$ ms and total model inference latency rises from $157.86$ to $167.11$ ms ($1.06\times$). \method{} uses one proposal and one backward pass, without a world model or candidate rollouts.

\subsection{From Local Ascent to Policy Improvement}
Equation~\ref{eq:local_ascent_bound} guarantees centralized critic-value improvement for a sufficiently small \method{} step, and paired rollouts confirm policy improvement across both benchmarks. \method{} raises Table~\ref{tab:main} Frozen reference on $20/24$ settings.

\paragraph{Beyond the generative backbone.}
\method{} needs only a frozen joint policy and a centralized behavior critic; the backbone need not be generative. On three separate settings with a frozen behavior-cloning MLP policy, one \method{} step improved every tested mean return. This transfer test shows that \method{} is a refinement operator for frozen joint policies, rather than a mechanism tied to the coordinated generative backbone.

\paragraph{Supplementary evidence}
 Appendix~A provides six checks that qualify the main operating-point results. Table~\ref{tab:smac-win} (in Appendix) pairs the shaped-return comparison with SMAC success: eight settings improve, two tie, and two decrease. Figure~\ref{fig:gain-quality} (in Appendix) reports above-chance critic ranking on $23/24$ settings and presents dataset quality and the two-factor headroom split as observational gain diagnostics. Table~\ref{tab:latency} (in Appendix) measures a $3.85$ ms mean cost for one critic backward pass and a $1.06\times$ model-side latency factor. Table~\ref{tab:mechanism} (in Appendix) reports positive full-joint-update gain on $22/24$ settings, with higher mean gain than the search and one-agent controls on most settings; its factorized comparison is mixed on continuous MPE. Table~\ref{tab:transfer-controls} (in Appendix) reports positive leave-one-task-out step transfer on most MPE settings and weaker transfer on SMAC, alongside nominal-radius control effects from separate batches. Figure~\ref{fig:dataset-keyframes} (in Appendix) displays the spatial coverage, joint pursuit, and focus-fire structures present in the evaluated tasks.

\section{Conclusion and Limitations}
This work introduced the \emph{coordination gap} perspective for frozen generative offline cooperative policies and \method{}, a test-time centralized value-guidance framework that closes recoverable local gaps without retraining. Its canonical action-space variant applies one normalized, projected joint-action gradient step with a formal local critic-ascent guarantee, one backward pass, and no world model or rollouts. Across $24$ MPE and SMAC settings, \method{} improves the equal-weighted mean setting-wise relative return by $9.1\%$ over the frozen backbone, while increasing model-side latency by only $1.06\times$ on average. The post-generation variant directly refines the decoded action with the behavior critic and avoids the separate normalized-coordinate critic required by trajectory-space injection. \method{} therefore restores test-time joint-action deliberation to frozen offline multi-agent policies without policy retraining.

\paragraph{Limitations.}
\method{} assumes centralized access to joint observations and actions and a critic that reliably ranks nearby joint actions. Its benefit is limited when the frozen action has little local headroom, and inaccurate critic ranking will reduce realized return. The useful step size depends on the task and action representation: weaker LOTO transfer on SMAC motivates domain-specific step selection. The experiments cover two benchmark families and selected operating points; evaluation on larger teams, stronger distribution shifts, and decentralized execution remains future work.

\raggedbottom
\clearpage
\bibliographystyle{plainnat}
\bibliography{G2MAF}

@article{zou2026coflow,
  title   = {CoFlow: Coordinated Few-Step Flow for Offline Multi-Agent Decision Making},
  author  = {Zou, Guowei and Wang, Haitao and Zhang, Beiwen and Zhang, Boning and Wu, Hejun},
  journal = {arXiv preprint arXiv:2605.01457},
  year    = {2026}
}

@inproceedings{fu2025ins,
  title={INS: Interaction-Aware Synthesis to Enhance Offline Multi-Agent Reinforcement Learning},
  author={Fu, Yuqian and Zhu, Yuanheng and Zhao, Jian and Chai, Jiajun and Zhao, Dongbin},
  booktitle={International Conference on Learning Representations},
  year={2025}
}

@article{qiao2025score,
  title={Offline Multi-Agent Reinforcement Learning via Score Decomposition},
  author={Qiao, Dan and Li, Wenhao and Yang, Shanchao and Zha, Hongyuan and Wang, Baoxiang},
  journal={arXiv preprint arXiv:2505.05968},
  year={2025}
}

@article{formanek2026data,
  title={Putting Data at the Centre of Offline Multi-Agent Reinforcement Learning},
  author={Formanek, Juan Claude and Beyers, Louise and Tilbury, Callum Rhys and Shock, Jonathan Phillip and Pretorius, Arnu},
  journal={Journal of Data-Centric Machine Learning Research},
  volume={3},
  number={11},
  pages={1--24},
  year={2026},
  url={https://openreview.net/forum?id=Rp6H7FKkpf}
}

@article{bagatella2025gcttt,
  title={Test-Time Offline Reinforcement Learning on Goal-Related Experience},
  author={Bagatella, Marco and Albaba, Mert and H{\"u}botter, Jonas and Martius, Georg and Krause, Andreas},
  journal={arXiv preprint arXiv:2507.18809},
  year={2025}
}

@article{alles2025flowq,
  title={FlowQ: Energy-Guided Flow Policies for Offline Reinforcement Learning},
  author={Alles, Marvin and Chen, Nutan and van der Smagt, Patrick and Cseke, Botond},
  journal={arXiv preprint arXiv:2505.14139},
  year={2025}
}

@article{tiofack2025gfp,
  title={Guided Flow Policy: Learning from High-Value Actions in Offline Reinforcement Learning},
  author={Tiofack, Franki Nguimatsia and Le Hellard, Th{\'e}otime and Schramm, Fabian and Perrin-Gilbert, Nicolas and Carpentier, Justin},
  journal={arXiv preprint arXiv:2512.03973},
  year={2025}
}

@article{chen2025onestep,
  title={One-Step Flow Policy Mirror Descent},
  author={Chen, Tianyi and Ma, Haitong and Li, Na and Wang, Kai and Dai, Bo},
  journal={arXiv preprint arXiv:2507.23675},
  year={2025}
}

@article{espinosa2025shortcut,
  title={Scaling Offline Reinforcement Learning via Efficient and Expressive Shortcut Models},
  author={Espinosa-Dice, Nicolas and Zhang, Yiyi and Chen, Yiding and Guo, Bradley and Oertell, Owen and Swamy, Gokul and Brantley, Kiante and Sun, Wen},
  journal={arXiv preprint arXiv:2505.22866},
  year={2025}
}

@article{frans2025guidance,
  title={Diffusion Guidance Is a Controllable Policy Improvement Operator},
  author={Frans, Kevin and Park, Seohong and Abbeel, Pieter and Levine, Sergey},
  journal={arXiv preprint arXiv:2505.23458},
  year={2025}
}

@inproceedings{li2026adjoint,
  title={Q-Learning with Adjoint Matching},
  author={Li, Qiyang and Levine, Sergey},
  booktitle={International Conference on Learning Representations},
  year={2026},
  url={https://openreview.net/forum?id=vd4eNAdtO6}
}

@inproceedings{mawam2027,
  title     = {{MA-WAM}: Multi-Agent World-Action Model for Test-Time Planning},
  author    = {Anonymous},
  booktitle = {Under review},
  year      = {2027},
  note      = {Companion submission}
}

@inproceedings{janner2022diffuser,
  title     = {Planning with Diffusion for Flexible Behavior Synthesis},
  author    = {Janner, Michael and Du, Yilun and Tenenbaum, Joshua B. and Levine, Sergey},
  booktitle = {International Conference on Machine Learning (ICML)},
  year      = {2022}
}

@inproceedings{ajay2023dd,
  title     = {Is Conditional Generative Modeling All You Need for Decision-Making?},
  author    = {Ajay, Anurag and Du, Yilun and Gupta, Abhi and Tenenbaum, Joshua B. and Jaakkola, Tommi S. and Agrawal, Pulkit},
  booktitle = {International Conference on Learning Representations (ICLR)},
  year      = {2023}
}

@inproceedings{zhu2023madiff,
  title     = {MADiff: Offline Multi-agent Learning with Diffusion Models},
  author    = {Zhu, Zhengbang and Liu, Minghuan and Mao, Liyuan and Kang, Bingyi and Xu, Minkai and Yu, Yong and Ermon, Stefano and Zhang, Weinan},
  booktitle = {Advances in Neural Information Processing Systems (NeurIPS)},
  year      = {2024}
}

@inproceedings{pan2022omar,
  title     = {Plan Better Amid Conservatism: Offline Multi-Agent Reinforcement Learning with Actor Rectification},
  author    = {Pan, Ling and Huang, Longbo and Ma, Tengyu and Xu, Huazhe},
  booktitle = {International Conference on Machine Learning (ICML)},
  year      = {2022}
}

@inproceedings{formanek2023ogmarl,
  title   = {Off-the-Grid MARL: Datasets and Baselines for Offline Multi-Agent Reinforcement Learning},
  author  = {Formanek, Claude and Jeewa, Asad and Shock, Jonathan and Pretorius, Arnu},
  booktitle = {Proceedings of the 22nd International Conference on Autonomous Agents and Multiagent Systems},
  pages = {2442--2444},
  publisher = {IFAAMAS},
  year = {2023}
}

@inproceedings{rashid2018qmix,
  title     = {QMIX: Monotonic Value Function Factorisation for Deep Multi-Agent Reinforcement Learning},
  author    = {Rashid, Tabish and Samvelyan, Mikayel and Schroeder de Witt, Christian and Farquhar, Gregory and Foerster, Jakob and Whiteson, Shimon},
  booktitle = {International Conference on Machine Learning (ICML)},
  year      = {2018}
}

@inproceedings{lowe2017maddpg,
  title     = {Multi-Agent Actor-Critic for Mixed Cooperative-Competitive Environments},
  author    = {Lowe, Ryan and Wu, Yi and Tamar, Aviv and Harb, Jean and Abbeel, Pieter and Mordatch, Igor},
  booktitle = {Advances in Neural Information Processing Systems (NeurIPS)},
  year      = {2017}
}

@inproceedings{yang2021believe,
  title={Believe what you see: Implicit constraint approach for offline multi-agent reinforcement learning},
  author={Yang, Yiqin and Ma, Xiaoteng and Li, Chenghao and Zheng, Zewu and Zhang, Qiyuan and Huang, Gao and Yang, Jun and Zhao, Qianchuan},
  booktitle={Advances in Neural Information Processing Systems},
  volume={34},
  pages={10299--10312},
  year={2021}
}

@inproceedings{kumar2020conservative,
  title={Conservative q-learning for offline reinforcement learning},
  author={Kumar, Aviral and Zhou, Aurick and Tucker, George and Levine, Sergey},
  booktitle={Advances in Neural Information Processing Systems},
  volume={33},
  pages={1179--1191},
  year={2020}
}

@inproceedings{fujimoto2021minimalist,
  title={A Minimalist Approach to Offline Reinforcement Learning},
  author={Fujimoto, Scott and Gu, Shixiang Shane},
  booktitle={Advances in Neural Information Processing Systems},
  volume={34},
  pages={20132--20145},
  year={2021}
}

@inproceedings{li2025dof,
  title={{DoF}: A Diffusion Factorization Framework for Offline Multi-Agent Reinforcement Learning},
  author={Li, Chao and Deng, Ziwei and Lin, Chenxing and Chen, Wenqi and Fu, Yongquan and Liu, Weiquan and Wen, Chenglu and Wang, Cheng and Shen, Siqi},
  booktitle={International Conference on Learning Representations},
  year={2025}
}

@inproceedings{chen2023sfbc,
  title={Offline reinforcement learning via high-fidelity generative behavior modeling},
  author={Chen, Huayu and Lu, Cheng and Ying, Chengyang and Su, Hang and Zhu, Jun},
  booktitle={International Conference on Learning Representations},
  year={2023}
}

@inproceedings{lee2025macflow,
  title={Multi-agent Coordination via Flow Matching},
  author={Lee, Dongsu and Lee, Daehee and Zhang, Amy},
  booktitle={International Conference on Learning Representations},
  year={2026}
}

@article{pomerleau1991efficient,
  title={Efficient Training of Artificial Neural Networks for Autonomous Navigation},
  author={Pomerleau, Dean A},
  journal={Neural Computation},
  volume={3},
  number={1},
  pages={88--97},
  year={1991}
}

@article{meng2021offline,
  title={Offline Pre-trained Multi-agent Decision Transformer},
  author={Meng, Linghui and Wen, Muning and Le, Chenyang and Li, Xiyun and Xing, Dengpeng and Zhang, Weinan and Wen, Ying and Zhang, Haifeng and Wang, Jun and Yang, Yaodong and Xu, Bo},
  journal={Machine Intelligence Research},
  volume={20},
  pages={233--248},
  year={2023},
  doi={10.1007/s11633-022-1383-7}
}

@article{li2026dom2,
  title={Improving Generalization and Data Efficiency with Diffusion in Offline Multi-agent Reinforcement Learning},
  author={Li, Zhuoran and Pan, Ling and Huang, Jiatai and Huang, Longbo},
  journal={Transactions on Machine Learning Research},
  year={2026}
}

@article{pang2026vgm2p,
  title={Value-Guidance MeanFlow for Offline Multi-Agent Reinforcement Learning},
  author={Pang, Teng and Dong, Zhiqiang and Zhang, Yan and Xu, Rongjian and Wu, Guoqiang and Yin, Yilong},
  journal={arXiv preprint arXiv:2604.08174},
  year={2026}
}

@inproceedings{zhan2025exploiting,
  title={Exploiting Structure in Offline Multi-Agent {RL}: The Benefits of Low Interaction Rank},
  author={Zhan, Wenhao and Fujimoto, Scott and Zhu, Zheqing and Lee, Jason D. and Jiang, Daniel R. and Efroni, Yonathan},
  booktitle={International Conference on Learning Representations},
  year={2025}
}

@inproceedings{liu2025inspo,
  title={{InSPO}: Offline Multi-Agent Reinforcement Learning via In-Sample Sequential Policy Optimization},
  author={Liu, Zongkai and Lin, Qian and Yu, Chao and Wu, Xiawei and Liang, Yile and Li, Donghui and Ding, Xuetao},
  booktitle={AAAI Conference on Artificial Intelligence},
  year={2025}
}

@inproceedings{park2025flowql,
  title={Flow {Q}-Learning},
  author={Park, Seohong and Li, Qiyang and Levine, Sergey},
  booktitle={International Conference on Machine Learning},
  year={2025}
}

@inproceedings{zhang2025efm,
  title={Energy-Weighted Flow Matching for Offline Reinforcement Learning},
  author={Zhang, Shiyuan and Zhang, Weitong and Gu, Quanquan},
  booktitle={International Conference on Learning Representations},
  year={2025}
}

@inproceedings{yuan2025madits,
  title={{MADiTS}: Efficient Multi-Agent Offline Coordination via Diffusion-Based Trajectory Stitching},
  author={Yuan, Lei and Bian, Yuqi and Li, Lihe and Zhang, Ziqian and Guan, Cong and Yu, Yang},
  booktitle={International Conference on Learning Representations},
  year={2025}
}

@article{li2025om2p,
  title={{OM2P}: Offline Multi-Agent Mean-Flow Policy},
  author={Li, Zhuoran and Wang, Xun and Zhong, Hai and Huang, Longbo},
  journal={arXiv preprint arXiv:2508.06269},
  year={2025}
}

@inproceedings{jang2025q,
  title={{Q}-Guided Flow {Q}-Learning},
  author={Jang, Yejun and Nam, Hong Chul and Park, Jeong Min and Bae, Gimin and Kwon, Hyun},
  booktitle={CoRL 2025 Workshop RemembeRL},
  year={2025}
}

@inproceedings{xu2026vgf,
  title={Reinforcement Learning via Value Gradient Flow},
  author={Xu, Haoran and Hu, Kaiwen and Sojoudi, Somayeh and Zhang, Amy},
  booktitle={International Conference on Learning Representations},
  year={2026}
}

@book{oliehoek2016concise,
  title={A Concise Introduction to Decentralized {POMDPs}},
  author={Oliehoek, Frans A. and Amato, Christopher},
  year={2016},
  publisher={Springer}
}

\appendix
\onecolumn
\raggedbottom

\section*{Appendix}
\subsection*{Contents}
\begingroup
\small
\setlength{\tabcolsep}{0pt}
\renewcommand{\arraystretch}{1.08}
\begin{tabular}{@{}p{\textwidth}@{}}
    \textbf{ Appendix section} \\
    \midrule
    \textbf{ Appendix~\ref{app:experiments}\quad Experimental Supplement} \\
    \quad \ref{app:q1-supp}\quad Q1: Gains Require Headroom and Reliable Value Ranking \\
    \quad \ref{app:q2-supp}\quad Q2: Value-Injection Locations in Few-Step Multi-Agent Flow \\
    \quad \ref{app:q3-supp}\quad Q3: Discrete  Actions, Step Size, and Cost \\
    \quad \ref{app:additional-supp}\quad  dditional Experiments Outside Q1 to Q3 \\
    \addlinespace[0.25em]
    \textbf{ Appendix~\ref{app:theory}\quad  Algorithm and Mathematical Derivations} \\
    \quad \ref{app:formal-setting}\quad Formal Multi-Agent Setting and Coordination Gap \\
    \quad \ref{app:critic-training}\quad Centralized Behavior Critic Training \\
    \quad \ref{app:algorithm}\quad Test-Time Joint- ction Refinement  Algorithm \\
    \quad \ref{app:discrete-derivation}\quad Discrete Joint  Actions \\
    \quad \ref{app:local-tilt}\quad Local Centralized Value-Tilt Derivation \\
    \quad \ref{app:cgb}\quad Coordinated Generative Backbone \\
    \quad \ref{app:injection-variants}\quad Value-Injection Variants for Coordinated Generators \\
    \bottomrule
\end{tabular}
\endgroup

\section{ Appendix  : Experimental Supplement}
\label{app:experiments}
Sections~1, 2, and 3 provide detailed results for Q1, Q2, and Q3, respectively. Section~4 separately reports step-size transfer, mechanism and structure controls, perturbation controls, and qualitative benchmark context. Each quantitative experiment states the question being tested, fixes the comparison protocol, defines the reported metrics and presentation conventions, reports the observed result, and limits the conclusion to that protocol.

\paragraph{Evaluation protocol.} The only \method{} hyperparameter is the step size $\eta$, swept independently for each setting: continuous $\eta\!\in\![0.01,0.1]$ with unit-normalized gradients and discrete $\eta\!\in\![2,50]$. Discrete steps are numerically larger because the selected action changes only after masked logits cross. Table~\ref{tab:main} reports the best value from each setting's grid, while Figure~\ref{fig:cont-step-sweeps} reports the corresponding sensitivity curves. The signed gain $\Delta_R$ is measured in the displayed benchmark metric. Relative gain $\delta_{\mathrm{rel}}$ is therefore compared only within a benchmark, because its value depends on that metric's reward origin.

\paragraph{Critic implementation.} For each setting, one centralized behavior critic $Q_\phi$ (Eq.~\ref{eq:fqe}) is fit on the \emph{same} offline dataset used to train the policy. It is a $4$-layer multilayer perceptron (MLP) with $512$ units per layer, LayerNorm, and sigmoid linear unit (SiLU) activations, operating on concatenated joint observations and actions. We train it for $20$k TD-regression steps with the  damW optimizer, $\gamma\!=\!0.99$, and a soft-updated target ($\tau\!=\!0.005$). For discrete domains, the action input concatenates per-agent one-hot vectors, which keeps $Q_\phi$ differentiable in the relaxed action probabilities. The target network $\bar Q$ forms TD targets only; \method{} differentiates the online critic $Q_\phi$ and trains no actor.

\subsection{Q1: Gains Require Headroom and Reliable Value Ranking}
\label{app:q1-supp}
\subsubsection{Main-Table Reporting Conventions}
Table~\ref{tab:main} reports MPE on the OMAR-normalized scale and SMAC as shaped episodic return. ``Data'' is the offline-dataset mean, and ``Diff'' abbreviates MADiff. Each \method{} point estimate is the selected operating point described in the main-text protocol. Its $\pm$ value is the population standard deviation across $K\in\{1,2,3,4,5\}$ in a separate sweep; MPE standard deviations use the same OMAR normalization as the point estimate. Published baselines retain the uncertainty convention of their source paper. Bold and underline identify the best and second-best method value in each row, excluding Data. Figure~\ref{fig:main-gains} uses the same selected operating points for the paired Frozen comparison.

\paragraph{Does \method{} improve a frozen policy, and are larger gains associated with recoverable local headroom?}
Q1 tests two linked claims: the post-generation update should improve more frozen operating points than it harms, and the largest gains should occur when the frozen proposal leaves nearby value headroom. The analysis treats headroom as a function of dataset quality, task dynamics, and policy saturation.

\emph{Experimental design.} Figure~\ref{fig:main-gains} reports paired differences between \method{} and Frozen under the fixed operating-point protocol. Table~\ref{tab:smac-win} separately measures SMAC success because shaped return and episode victory are not equivalent outcomes. Figure~\ref{fig:gain-quality}(a,b) groups the same operating-point gains by dataset quality for MPE and SMAC. The later reliability experiment uses held-out simulator outcomes rather than the FQE training loss.

\emph{Metric and comparison conventions.} In Table~\ref{tab:smac-win}, $\Delta_{\rm win}$ is the percentage-point success-rate change; $\Delta_R$ is the shaped-return gain. Figures~\ref{fig:gain-quality}(a,b) plot task-level gain against dataset quality; each line uses its own benchmark's available quality levels.

\begin{table}[H]
\centering
{\papertablesize
\setlength{\tabcolsep}{6pt}
\begin{tabular*}{\textwidth}{@{\extracolsep{\fill}}llccc@{\hskip 2em}llccc@{}}
\toprule
Map & Quality & Frozen \% & \method{} \% & $\Delta_{\rm win}$
& Map & Quality & Frozen \% & \method{} \% & $\Delta_{\rm win}$ \\
\midrule
3m   & Good   & 95.6 & 95.6 & $0.0$
& 5m\_vs\_6m & Good   & 74.4 & 78.9 & $\textbf{+4.4}$ \\
     & Medium & 66.7 & 50.0 & $-16.7$
&             & Medium & 68.9 & 77.8 & $\textbf{+8.9}$ \\
     & Poor   & 22.2 & 43.3 & $\textbf{+21.1}$
&             & Poor   & 13.3 & 25.6 & $\textbf{+12.2}$ \\
2s3z & Good   & 97.8 & 95.6 & $-2.2$
& 8m          & Good   & 96.7 & 98.9 & $\textbf{+2.2}$ \\
     & Medium & 65.6 & 76.7 & $\textbf{+11.1}$
&             & Medium & 76.7 & 88.9 & $\textbf{+12.2}$ \\
     & Poor   & 0.0 & 4.4 & $\textbf{+4.4}$
&             & Poor   & 0.0 & 0.0 & $0.0$ \\
\bottomrule
\end{tabular*}}
\caption{\textbf{SMAC success rate} ($\uparrow$, our policy only): Frozen backbone vs.\ \method{} at each method's selected denoising-step count. Success and shaped episode return measure different outcomes; success is therefore reported separately. Here $\Delta_{\rm win}=W_{\rm \method{}}-W_{\rm Frozen}$ is the percentage-point change in success rate; $\Delta_R$ is the shaped-return gain.}
\label{tab:smac-win}
\end{table}

\noindent\emph{Result analysis.} Table~\ref{tab:smac-win} reports higher success on eight SMAC settings, two ties, and two decreases. Figure~\ref{fig:gain-scatter}(a) groups Table~\ref{tab:main} point estimates by dataset quality. The lowest-quality endpoint has the largest benchmark-level mean relative gain on MPE ($+32.5\%$) and SMAC ($+15.0\%$). These means summarize the evaluated operating points within each benchmark.

\noindent\emph{Conclusion.} The fixed operating-point comparison shows that \method{} improves a majority of the evaluated settings, with gain magnitude varying across domains. Lower-quality MPE and SMAC settings usually show larger gains, which is consistent with greater recoverable headroom. The next experiment measures realized local headroom and critic ranking reliability directly.

\subsubsection{Refinement Locality Relative to Empirical Support}
\paragraph{Does the update remain local on the scale of behavior-supported actions?}
For each of the $12$ continuous MPE settings, we apply one \method{} step to dataset joint actions and divide its displacement by the nearest-neighbor action distance at matched behavior states. The evaluated update remains below the characteristic nearest-neighbor action scale of the offline data and exhibits the local value ascent described by Eq.~\ref{eq:local_ascent_bound}.

\subsubsection{Direct Critic Reliability and a Two-Factor Headroom Test}

\paragraph{Does the behavior critic rank realized local outcomes, and does gain require both headroom and reliable ranking?}
The TD loss in Eq.~\ref{eq:fqe} measures regression fit on logged transitions. The ranking property required by \method{} is therefore evaluated separately against realized returns of executable actions. The experiment then tests the relation of headroom and reliability to large gains.

\emph{Experimental design.} We first run the frozen $K{=}5$ policy in held-out simulator episodes: $20$ per MPE setting and $10$ per SMAC setting. At every timestep with at least eight rewards remaining, we compare $Q_\phi(\bm o_t,\bm a_t)$ with the fixed-window target $G_t^{(8)}=\sum_{h=0}^{7}\gamma^h r_{t+h}$. The common horizon removes remaining-episode-length variation. Pairwise accuracy is computed within each episode and then averaged, giving every episode equal weight.

\begin{figure*}[!t]
\centering
\includegraphics[width=\textwidth]{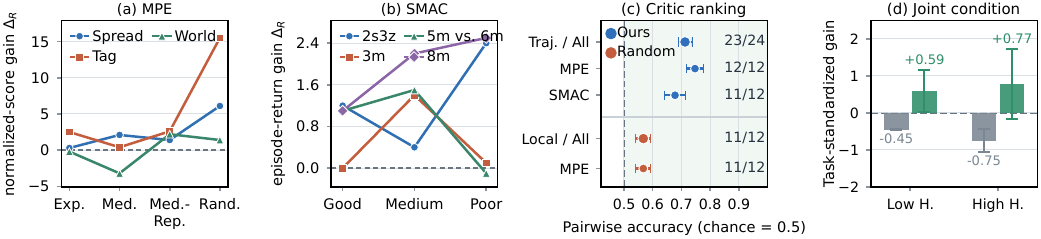}
\caption{\textbf{Quality and refinement diagnostics.} Panels (a) and (b) show task-level Table~\ref{tab:main} gain across dataset qualities for MPE and SMAC, respectively; MPE uses normalized-score gain and SMAC uses shaped episodic-return gain. (c) Critic pairwise accuracy against simulator-realized returns. Trajectory rows use frozen-policy trajectories; local rows use the Frozen action and $14$ random proposals with matched nominal pre-projection radius from $64$ restored states per MPE setting. Points and error bars are means $\pm1$ standard error across settings; right-hand labels count settings above chance. (d) Mean task-standardized MPE gain after splitting realized headroom at zero and independent local reliability at its median; gray and green denote low and high reliability. Labels report the cell mean.}
\label{fig:gain-quality}
\label{fig:gain-scatter}
\label{fig:critic-reliability}
\label{fig:headroom-twofactor}
\end{figure*}

The stricter local test restores the same simulator state before evaluating the Frozen action and $14$ random actions. Each random proposal has nominal pre-projection radius $\|\tilde{\bm a}-\bm a_\theta\|_2$, the realized displacement of the \method{} action, and is then projected to the same action box. Let $G_H(\bm a)$ denote the realized $H{=}8$ return under their shared continuation. The random pool defines the empirical, \method{}-independent headroom
\begin{equation}
\widehat{\mathcal G}^{\rm emp}_{H}
=\left[\max_{1\le j\le14}G_H(\bm a_j^{\rm rand})
-G_H(\bm a_\theta)\right]_+,
\label{eq:empirical-headroom}
\end{equation}
where box projection makes $\|\bm a_j^{\rm rand}-\bm a_\theta\|_2$ smaller than the nominal radius. Local reliability is the same-state pairwise accuracy on $\{\bm a_\theta,\bm a_1^{\rm rand},\ldots,\bm a_{14}^{\rm rand}\}$. Neither factor includes the \method{} action or its realized gain.

\emph{Metric and comparison conventions.} Figure~\ref{fig:critic-reliability}(c) compares critic pairwise accuracy with the $0.5$ chance level; the right-hand labels count settings above chance. Figure~\ref{fig:headroom-twofactor}(d) splits empirical headroom at zero and local reliability at its median. Each bar is the mean task-standardized Table~\ref{tab:main} gain. This standardization is used solely for the descriptive cross-task grouping.

\noindent\emph{Result analysis.} The behavior-trajectory test exceeds chance in all twelve MPE settings and eleven of twelve SMAC settings; the exception is the near-saturated \texttt{3m-Good} setting. Using Table~\ref{tab:main} gains in Figure~\ref{fig:headroom-twofactor}(d), the low-headroom/low-reliability, low-headroom/high-reliability, high-headroom/low-reliability, and high-headroom/high-reliability cells have mean standardized gains of $-0.45$, $+0.59$, $-0.75$, and $+0.77$, respectively. Their positive-setting counts are $2/2$, $4/4$, $2/4$, and $2/2$.

\noindent\emph{Conclusion.} Held-out simulator outcomes show that the critic usually ranks the tested neighborhoods better than chance. The largest Table~\ref{tab:main} gains are distributed across the four headroom--reliability cells. This diagnostic therefore evaluates critic ranking and describes the boundary of the headroom hypothesis.

\FloatBarrier

\FloatBarrier
\subsection{Q2: \method{} Value-Injection Locations in Few-Step Multi-Agent Flow}
\label{app:q2-supp}

\paragraph{Does the point of value-gradient injection change the realized return gain?}
Q2 compares three \method{} variants: applying value guidance throughout generation, only at the final flow step, or after action decoding. It tests whether the injection location produces the same ranking across tasks and whether post-generation refinement is a uniformly best return optimizer or a deployment choice with a different implementation interface.

\emph{Experimental design.} All $12$ continuous MPE settings use the same frozen coordinated backbone. ``All'' applies the trajectory-space \method{} update at every flow step and ``Final'' applies it only at the last flow step; both retain the best gain from the independent fixed-$K{=}5$ guidance-scale rerun. ``Post'' is the action-space \method{} variant and therefore reuses Table~\ref{tab:main} Frozen and refined point estimates. All three MPE gains are displayed on Table~\ref{tab:main} OMAR-normalized scale. SMAC is excluded because its discrete update operates on masked logits before the final $\arg\max$.

\emph{Variant and comparison conventions.} All and Final are independent trajectory-space \method{} effect estimates, whereas Post is the canonical action-space \method{} subtraction. Figure~\ref{fig:q2-q3-dashboard} presents the three gains by task and dataset quality. The separate evaluation batches make this ordering descriptive.

\noindent\emph{Result analysis.} Across the $12$ MPE settings, all-step trajectory-space \method{} has the largest displayed gain on $8$ settings and is positive on all settings. Final-step trajectory-space \method{} is largest on $1$ setting and is positive on all settings. The canonical post-generation action-space \method{} gain is largest on $3$ settings and is positive on $10$.

\noindent\emph{Conclusion.} The separate batches yield descriptive values for the three injection points. Post-generation action-space \method{} is the deployment choice because it acts on the executable action and reuses the behavior critic; the all-step and final-step trajectory-space variants require a separate normalized-coordinate critic.

\FloatBarrier
\subsection{Q3: Discrete  Actions, Step Size, and Cost}
\label{app:q3-supp}

\paragraph{Does one local-gradient interface handle continuous actions and discrete decisions at practical inference cost?}
Q3 tests two claims. First, $\eta$ should control update locality in both action space and masked-logit space, although the useful numerical scale may differ because a discrete $\arg\max$ changes only after logits cross. Second, one critic backward pass should add substantially less deployment machinery than candidate rollout through a world model.

\emph{Experimental design.} Figure~\ref{fig:cont-step-sweeps}(a) evaluates the complete grid $\eta\in\{0,0.01,0.03,0.05,0.1,0.2,0.3,0.5\}$ on all $12$ continuous MPE settings with fixed $K{=}5$. Panel~(b) sweeps masked-logit steps $\{0,2,10,50\}$ on all $12$ SMAC settings. These matched sensitivity reruns retain Table~\ref{tab:main} operating points and compare each step with its own Base row. Table~\ref{tab:latency} times the Frozen backbone, the gradient computation, and the complete refined decision on every setting.

\emph{Metric and plotting conventions.} Every curve reports $\Delta_R=R_\eta-R_{\mathrm{Base}}$; equal horizontal spacing identifies the tested values, while their numerical magnitudes are given by the axis labels. Rings in panel~(b) mark the best tested logit step for each setting. In Table~\ref{tab:latency}, Grad is the direct backward-pass time, while $\times$ divides the measured complete \method{} latency by the measured Base latency for that row.

\FloatBarrier

\noindent\emph{Result analysis.} In the continuous grid, the within-sweep maximizing step varies across settings.  At $\eta=0.5$, \texttt{Spread-Random} and \texttt{Tag-Random} reach gains of $+16.2$ and $+35.6$ normalized points. The SMAC optima vary across maps and qualities and commonly require steps of order $10$ or larger. Across all $24$ settings, the gradient computation averages $3.85$ ms, and mean complete model-side latency changes from $157.86$ to $167.11$ ms, a factor of $1.06\times$.

\noindent\emph{Conclusion.} The same normalized-gradient interface applies to continuous actions and discrete masked logits, with action-representation-specific numerical step scales. The sweeps support treating $\eta$ as an action-representation-dependent locality parameter. The measured backward pass averages $3.85$ ms over these settings; the timing covers model-side computation only.

\begin{table*}[t]
\centering
{\papertablesize
\setlength{\tabcolsep}{0pt}
\begin{tabular*}{\textwidth}{@{\extracolsep{\fill}}llrrrr@{\hspace{1.2em}}llrrrr@{}}
\toprule
Task & Quality & \shortstack{Base\\(ms)} & \shortstack{Grad\\(ms)} & \shortstack{\method{}\\(ms)} & $\times$
& Task & Quality & \shortstack{Base\\(ms)} & \shortstack{Grad\\(ms)} & \shortstack{\method{}\\(ms)} & $\times$ \\
\midrule
\multicolumn{6}{l}{\textit{MPE}}
& \multicolumn{6}{l}{\textit{SMAC}} \\
Spread & Expert    & 142.1 & 3.0 & 145.1 & 1.02 & 3m         & Good   & 189.9 & 3.4 & 179.5 & 0.94 \\
       & MedReplay & 137.4 & 3.0 & 142.7 & 1.04 &            & Medium & 175.7 & 3.9 & 188.5 & 1.07 \\
       & Medium    & 140.5 & 3.2 & 149.1 & 1.06 &            & Poor   & 221.2 & 6.8 & 233.2 & 1.05 \\
       & Random    & 137.3 & 2.9 & 143.9 & 1.05 & 2s3z       & Good   & 153.8 & 4.3 & 181.9 & 1.18 \\
Tag    & Expert    & 138.9 & 3.1 & 150.7 & 1.08 &            & Medium & 164.2 & 3.7 & 162.4 & 0.99 \\
       & MedReplay & 135.6 & 3.3 & 147.3 & 1.09 &            & Poor   & 143.7 & 3.4 & 165.4 & 1.15 \\
       & Medium    & 164.5 & 5.9 & 168.1 & 1.02 & 5m\_vs\_6m & Good   & 181.3 & 3.6 & 178.7 & 0.99 \\
       & Random    & 141.9 & 3.2 & 144.1 & 1.02 &            & Medium & 171.3 & 3.6 & 163.8 & 0.96 \\
World  & Expert    & 136.7 & 2.7 & 139.9 & 1.02 &            & Poor   & 162.9 & 3.3 & 176.8 & 1.09 \\
       & MedReplay & 141.0 & 3.6 & 148.0 & 1.05 & 8m         & Good   & 165.9 & 3.5 & 179.4 & 1.08 \\
       & Medium    & 139.2 & 3.2 & 144.6 & 1.04 &            & Medium & 176.9 & 3.8 & 177.7 & 1.00 \\
       & Random    & 155.5 & 5.9 & 156.8 & 1.01 &            & Poor   & 171.3 & 6.3 & 242.9 & 1.42 \\
\midrule
\multicolumn{2}{l}{\textbf{All (24-setting mean)}} & 157.9 & 3.9 & 167.1 & 1.06
& \multicolumn{6}{l}{} \\
\bottomrule
\end{tabular*}}
\caption{Model-side inference latency in milliseconds per decision for every evaluated setting. Base is the frozen backbone; Grad is the direct \method{} critic-gradient step; \method{} is the full refined policy.}
\label{tab:latency}
\end{table*}

\noindent\emph{Latency takeaway.} Across the $24$ measured settings, the direct critic-gradient step adds $3.85$ ms on average and changes complete model-side decision latency from $157.86$ to $167.11$ ms ($1.06\times$).

\begin{table}[H]
\centering
{\papertablesize\textbf{(a) Structure and search controls}}\par\vspace{0.25em}
{\papertablesize
\renewcommand{\arraystretch}{0.96}
\setlength{\tabcolsep}{0pt}
\begin{tabular*}{\textwidth}{@{\extracolsep{\fill}}llrrrrr@{\hspace{0.8em}}llrrrrr@{}}
\toprule
Task & Quality & Frozen & \method{} & Best-$N$ & 1-agent & Factor.
& Task & Quality & Frozen & \method{} & Best-$N$ & 1-agent & Factor. \\
\midrule
\multicolumn{7}{l}{\emph{MPE}} & \multicolumn{7}{l}{\emph{SMAC}} \\
Spread & Expert    & 108.59 & $+0.24$ & $\mathbf{+1.19}$ & $+0.22$ & $+0.05$ & 3m          & Good   & 19.17 & $+0.01$ & $\mathbf{+0.03}$ & $+0.01$ & -- \\
       & MedReplay & 43.29 & $\mathbf{+1.72}$ & $-2.48$ & $-0.03$ & $+1.62$ &             & Medium & 15.12 & $-1.88$ & $-1.70$ & $\mathbf{+0.79}$ & -- \\
       & Medium    & 59.35 & $\mathbf{+3.50}$ & $-4.74$ & $+1.24$ & $+2.21$ &             & Poor   & 9.14 & $\mathbf{+1.72}$ & $+1.04$ & $-0.03$ & -- \\
       & Random    & 36.69 & $\mathbf{+6.84}$ & $+5.93$ & $+1.59$ & $+6.30$ & 2s3z        & Good   & 19.54 & $+0.32$ & $+0.26$ & $\mathbf{+0.46}$ & -- \\
Tag    & Expert    & 113.02 & $+6.46$ & $+7.45$ & $+5.05$ & $\mathbf{+10.57}$ &             & Medium & 17.08 & $\mathbf{+1.62}$ & $+0.75$ & $-0.06$ & -- \\
       & MedReplay & 60.90 & $+3.77$ & $-0.05$ & $+0.66$ & $\mathbf{+5.99}$ &             & Poor   & 9.87 & $\mathbf{+2.16}$ & $-0.72$ & $+0.57$ & -- \\
       & Medium    & 94.58 & $\mathbf{+5.24}$ & $+0.05$ & $-0.61$ & $+0.05$ & 5m\_vs\_6m & Good   & 16.97 & $\mathbf{+0.25}$ & $-0.66$ & $-0.05$ & -- \\
       & Random    & 20.24 & $+10.33$ & $+10.00$ & $+4.58$ & $\mathbf{+12.36}$ &             & Medium & 15.24 & $+2.32$ & $+1.64$ & $\mathbf{+2.97}$ & -- \\
World  & Expert    & 149.35 & $+2.50$ & $\mathbf{+3.59}$ & $+1.41$ & $+2.72$ &             & Poor   & 11.58 & $-1.22$ & $-0.12$ & $\mathbf{+0.06}$ & -- \\
       & MedReplay & 50.11 & $+3.26$ & $-0.33$ & $-0.33$ & $\mathbf{+6.20}$ & 8m          & Good   & 19.07 & $\mathbf{+0.63}$ & $+0.40$ & $-0.09$ & -- \\
       & Medium    & 112.72 & $\mathbf{+4.02}$ & $+2.17$ & $-0.54$ & $+1.63$ &             & Medium & 17.90 & $\mathbf{+1.06}$ & $+0.40$ & $-0.45$ & -- \\
       & Random    & 0.33 & $\mathbf{+0.98}$ & $+0.00$ & $+0.22$ & $+0.65$ &             & Poor   & 6.62 & $\mathbf{+2.18}$ & $+1.05$ & $+0.33$ & -- \\
\bottomrule
\end{tabular*}
}\par
\vspace{1.5em}
\noindent{\papertablesize\textbf{(b) Local-step diagnostics}}\par\vspace{0.25em}
{\papertablesize
\renewcommand{\arraystretch}{0.96}
\setlength{\tabcolsep}{0pt}
\begin{tabular*}{\textwidth}{@{\extracolsep{\fill}}llrrrr@{\hspace{1.2em}}llrrrr@{}}
\toprule
Task & Quality & Return & $\Delta_Q$ & Norm & Bound.\ frac.
& Task & Quality & Return & $\Delta_Q$ & Norm & Bound.\ frac. \\
\midrule
\multicolumn{6}{l}{\emph{MPE}} & \multicolumn{6}{l}{\emph{SMAC}} \\
Spread & Expert    & 108.84 & $+3.21$ & 0.030 & 0.017 & 3m          & Good   & 19.18 & $+0.03$ & 0.999 & -- \\
       & MedReplay & 45.02 & $+9.79$ & 0.098 & 0.030 &             & Medium & 13.24 & $+0.11$ & 9.076 & -- \\
       & Medium    & 62.85 & $+9.85$ & 0.098 & 0.028 &             & Poor   & 10.86 & $+0.25$ & 43.610 & -- \\
       & Random    & 43.51 & $+5.31$ & 0.100 & 0.002 & 2s3z        & Good   & 19.86 & $+0.32$ & 10.000 & -- \\
Tag    & Expert    & 119.53 & $+3.85$ & 0.047 & 0.081 &             & Medium & 18.70 & $+0.33$ & 1.941 & -- \\
       & MedReplay & 64.67 & $+54.19$ & 0.096 & 0.047 &             & Poor   & 12.04 & $+0.31$ & 27.311 & -- \\
       & Medium    & 99.81 & $+16.87$ & 0.087 & 0.129 & 5m\_vs\_6m & Good   & 17.22 & $+0.08$ & 1.908 & -- \\
       & Random    & 30.52 & $+2.51$ & 0.098 & 0.042 &             & Medium & 17.56 & $+0.16$ & 1.923 & -- \\
World  & Expert    & 151.85 & $+3.89$ & 0.082 & 0.149 &             & Poor   & 10.37 & $+0.17$ & 1.722 & -- \\
       & MedReplay & 53.37 & $+5.46$ & 0.089 & 0.097 & 8m          & Good   & 19.70 & $+0.54$ & 2.000 & -- \\
       & Medium    & 116.74 & $+13.02$ & 0.088 & 0.148 &             & Medium & 18.95 & $+2.79$ & 9.835 & -- \\
       & Random    & 1.41 & $+0.16$ & 0.028 & 0.148 &             & Poor   & 8.80 & $+1.76$ & 30.005 & -- \\
\bottomrule
\end{tabular*}
}
\caption{\textbf{Full-grid mechanism controls and local-step diagnostics over all $24$ settings.} (a) Paired mean performance gains for the full centralized update and the structure/search controls. (b) Critic-predicted value change and realized update geometry from the same rerun.}
\label{tab:mechanism}
\end{table}

\subsection{ .4  dditional Experiments Outside Q1 to Q3}
\label{app:additional-supp}
This section collects auxiliary experiments on the components of the local update, step-size transfer to held-out settings, matched nominal-radius perturbations, and the behaviors represented by benchmark trajectories.

\subsubsection{Mechanism and Structure Controls}

\paragraph{How does the centralized joint gradient compare with local search and partial or factorized updates?}
This experiment separates three properties of \method{}: using a gradient instead of sampling local directions, updating every agent rather than one agent, and evaluating the joint action with one centralized critic rather than independent per-agent critics.

\emph{Experimental design.} Frozen, the complete \method{}, Best-$N$, and 1-agent use the same evaluation protocol on all $24$ task-quality settings. Each update uses the setting-specific step selected by the step-size sweep. Best-$N$ draws $N{=}16$ unit-norm perturbations at the \method{} radius, scores them with the same centralized critic, and retains the highest-scoring candidate. The 1-agent control applies the centralized gradient to one agent only. Factorized trains decentralized behavior critics $Q_i(\bm o_i,\bm a_i)$ and ascends $\sum_iQ_i$ with the same step and action-set projection rule on all $12$ continuous MPE settings. The Factorized control is reported for MPE, whose continuous-action implementation admits this per-agent critic bundle.

\emph{Metric and table conventions.} In Table~\ref{tab:mechanism}(a), Frozen is the mean unrefined MPE OMAR-normalized score or SMAC shaped return. \method{}, Best-$N$, 1-agent, and Factor.\ are paired gains in the same metric. Best-$N$ selects the highest critic-scored action from $16$ normalized random perturbations at the \method{} radius; 1-agent applies the centralized critic gradient to one agent only; and Factor.\ ascends the sum of learned decentralized per-agent critics. Bold marks the largest method gain in each task-quality row. In Table~\ref{tab:mechanism}(b), Return is the MPE OMAR-normalized score or SMAC shaped return after the complete \method{} update; $\Delta_Q$ is the behavior critic's predicted before--after value increase; Norm is the realized update $\ell_2$ norm (action space for MPE and logit space for SMAC); and Bound.\ frac.\ is the fraction of MPE action coordinates that lie at a box boundary after projection. Table~\ref{tab:mechanism}(b) reports implementation diagnostics. `--' marks the MPE-only Factor.\ control in (a) and the inapplicable Bound.\ frac.\ for masked discrete actions in (b). The estimates summarize mean ordering within each row.

\FloatBarrier

\noindent\emph{Result analysis.} The \method{} gain is positive on $22/24$ settings. Its mean gain is higher than Best-$N$ on $18/24$ settings and higher than the 1-agent gain on $19/24$. The centralized update exceeds Factorized on $7/12$ continuous MPE settings, yielding a mixed comparison across tasks and qualities. In Table~\ref{tab:mechanism}(b), $\Delta_Q$ is positive on all $24$ settings. Continuous displacement follows the setting-specific radius, and SMAC updates remain legal after action re-masking.

\noindent\emph{Conclusion.} Across paired mean estimates, the full joint gradient improves more settings than sampled search or one-agent correction; the factorized comparison remains mixed on continuous tasks. Table~\ref{tab:mechanism}(b) verifies that the implemented step increases its own critic value while respecting the intended local geometry. Paired return evaluations provide the corresponding realized-return evidence.

\FloatBarrier

\begin{table}[!t]
\centering
{\papertablesize\textbf{(a) Non-oracle leave-one-task-out step-size transfer}}\par\vspace{0.25em}
{\papertablesize
\renewcommand{\arraystretch}{0.94}
\setlength{\tabcolsep}{0pt}
\begin{tabular*}{\textwidth}{@{\extracolsep{\fill}}llrrr@{\hspace{1.2em}}llrrr@{}}
\toprule
Task & Quality & \shortstack{Canonical\\$\Delta_R$} & \shortstack{LOTO\\$\eta$} & \shortstack{LOTO\\$\Delta_R$}
& Task & Quality & \shortstack{Canonical\\$\Delta_R$} & \shortstack{LOTO\\$\eta$} & \shortstack{LOTO\\$\Delta_R$} \\
\midrule
\multicolumn{5}{l}{\emph{MPE}} & \multicolumn{5}{l}{\emph{SMAC}} \\
Spread & Expert    & $+0.30$  & 0.1 & $-0.76$  & 3m          & Good   & $+0.00$ & 1  & $+0.00$ \\
       & MedReplay & $+1.40$  & 0.1 & $+2.12$  &             & Medium & $+1.40$ & 2  & $-1.75$ \\
       & Medium    & $+2.10$  & 0.1 & $+4.97$ &             & Poor   & $+0.10$ & 2  & $-1.61$ \\
       & Random    & $+6.10$ & 0.1 & $+7.62$ & 2s3z        & Good   & $+1.20$ & 2  & $+1.22$ \\
Tag    & Expert    & $+2.50$  & 0.1 & $+4.76$ &             & Medium & $+0.40$ & 2  & $+0.40$ \\
       & MedReplay & $+2.60$  & 0.1 & $+10.38$ &             & Poor   & $+2.40$ & 2  & $+1.21$ \\
       & Medium    & $+0.40$  & 0.1 & $+11.86$ & 5m\_vs\_6m & Good   & $+1.10$ & 2  & $+1.08$ \\
       & Random    & $+15.50$ & 0.1 & $+13.79$ &             & Medium & $+1.50$ & 10 & $-4.30$ \\
World  & Expert    & $-0.20$  & 0.1 & $+14.72$ &             & Poor   & $-0.10$ & 2  & $-0.14$ \\
       & MedReplay & $+2.20$  & 0.1 & $+13.14$ & 8m          & Good   & $+1.10$ & 2  & $+1.12$ \\
       & Medium    & $-3.20$  & 0.1 & $+13.60$ &             & Medium & $+2.20$ & 2  & $+1.95$ \\
       & Random    & $+1.40$  & 0.1 & $+0.24$  &             & Poor   & $+2.50$ & 2  & $+0.67$ \\
\cmidrule(r){1-5}\cmidrule(l){6-10}
\multicolumn{2}{l}{LOTO positive} & \multicolumn{3}{c}{11/12}
& \multicolumn{2}{l}{LOTO positive} & \multicolumn{3}{c}{7/12} \\
\bottomrule
\end{tabular*}
}\par
\vspace{1.5em}
\noindent{\papertablesize\textbf{(b) Matched nominal-radius perturbation controls}}\par\vspace{0.25em}
{\papertablesize
\renewcommand{\arraystretch}{0.94}
\setlength{\tabcolsep}{0pt}
\begin{tabular*}{\textwidth}{@{\extracolsep{\fill}}llrrrr@{\hspace{1.2em}}llrrrr@{}}
\toprule
Task & Quality & Base & \shortstack{\method{}\\$\Delta_R$} & \shortstack{Random\\$\Delta_R$} & \shortstack{Shuffled\\$\Delta_R$}
& Task & Quality & Base & \shortstack{\method{}\\$\Delta_R$} & \shortstack{Random\\$\Delta_R$} & \shortstack{Shuffled\\$\Delta_R$} \\
\midrule
\multicolumn{6}{l}{\textit{MPE (OMAR-normalized score)}}
& \multicolumn{6}{l}{\textit{SMAC (episode return)}} \\
Spread & Expert & 114.90 & $\mathbf{+0.30}$ & $-1.99$ & $-1.64$ & 3m & Good & 20.0 & $\mathbf{+0.0}$ & $-0.2$ & $-1.1$ \\
 & MedReplay & 45.91 & $\mathbf{+1.40}$ & $-0.27$ & $-0.08$ &  & Medium & 20.0 & $\mathbf{+1.4}$ & $-1.4$ & $-2.4$ \\
 & Medium & 57.60 & $\mathbf{+2.10}$ & $-8.30$ & $-3.42$ &  & Poor & 14.7 & $\mathbf{+0.1}$ & $-8.3$ & $-4.2$ \\
 & Random & 51.91 & $\mathbf{+6.09}$ & $+2.34$ & $-1.10$ & 2s3z & Good & 20.0 & $\mathbf{+1.2}$ & $+0.0$ & $-0.8$ \\
Tag & Expert & 137.22 & $+2.50$ & $\mathbf{+8.87}$ & $+5.42$ &  & Medium & 18.7 & $+0.4$ & $-0.3$ & $\mathbf{+1.0}$ \\
 & MedReplay & 70.90 & $+2.59$ & $+4.10$ & $\mathbf{+5.66}$ &  & Poor & 9.8 & $\mathbf{+2.4}$ & $-3.8$ & $+1.1$ \\
 & Medium & 112.31 & $\mathbf{+0.38}$ & $-5.85$ & $-5.75$ & 5m\_vs\_6m & Good & 18.3 & $\mathbf{+1.1}$ & $-0.2$ & $+1.1$ \\
 & Random & 30.61 & $\mathbf{+15.52}$ & $+3.35$ & $+3.68$ &  & Medium & 19.1 & $\mathbf{+1.5}$ & $+0.7$ & $-0.6$ \\
World & Expert & 156.63 & $-0.22$ & $+8.59$ & $\mathbf{+9.02}$ &  & Poor & 12.5 & $-0.1$ & $\mathbf{+1.0}$ & $-0.6$ \\
 & MedReplay & 43.59 & $+2.17$ & $\mathbf{+12.17}$ & $+9.46$ & 8m & Good & 20.0 & $\mathbf{+1.1}$ & $+0.3$ & $+0.7$ \\
 & Medium & 150.00 & $-3.15$ & $+7.93$ & $\mathbf{+12.83}$ &  & Medium & 17.5 & $\mathbf{+2.2}$ & $+1.7$ & $+1.4$ \\
 & Random & 3.59 & $\mathbf{+1.41}$ & $-0.22$ & $+0.87$ &  & Poor & 7.0 & $\mathbf{+2.5}$ & $-0.4$ & $+1.9$ \\
\bottomrule
\end{tabular*}}
\caption{\textbf{Full-grid non-oracle step-size transfer and perturbation controls over all $24$ settings.} (a) Canonical gain, the step selected from the other settings in the same domain, and the independently evaluated held-out gain. (b) Canonical Base and \method{} gain together with independently measured matched nominal-radius Random and Shuffled effects. MPE values use Table~\ref{tab:main} OMAR-normalized scale; SMAC values use shaped episodic return. Results across the two panels use different evaluation batches and are compared descriptively.}
\label{tab:transfer-controls}
\end{table}

\subsubsection{Non-Oracle Step-Size Transfer}

\paragraph{Does a step size selected without the held-out setting retain a positive gain?}
This experiment uses a stricter selection protocol: $\eta$ is selected before the held-out result is evaluated.

\emph{Experimental design.} For each held-out task-quality setting, LOTO selection chooses the single $\eta$ with the largest mean relative gain on the other settings from the same domain. The chosen value is then evaluated on the held-out setting. The complete-method reference is the canonical Table~\ref{tab:main} gain.

\emph{Metric and comparison conventions.} Table~\ref{tab:transfer-controls}(a) reports the canonical Table~\ref{tab:main} gain, the step selected only from the other settings in the same domain, and the independently evaluated held-out gain. The summary rows count positive LOTO gains. Canonical and LOTO values remain separate because they use different evaluation batches.

\noindent\emph{Result analysis.} LOTO gives positive held-out gains on $11/12$ MPE settings and $7/12$ SMAC settings, for $18/24$ overall. The canonical Table~\ref{tab:main} gains are positive on $10/12$ settings in each benchmark. The table presents these two evaluation batches side by side for reference.

\noindent\emph{Conclusion.}  A domain-level transferred step remains positive on most MPE settings but only seven of twelve SMAC settings. The observed transfer pattern favors action-representation-specific step sizes.

\subsubsection{Perturbation Controls}

\paragraph{Does gain come from the correctly conditioned critic direction rather than from perturbing the action by the same radius?}
 A positive local gain could arise because many nearby actions improve on the Frozen proposal, even if the critic direction carries no useful information. The random and shuffled-observation controls test this alternative explanation.

\emph{Experimental design.} Base and \method{} reproduce Table~\ref{tab:main} operating-point values on the same MPE OMAR-normalized and SMAC shaped-return scales. Random replaces the critic gradient with a random unit direction, and Shuffled computes a gradient after mismatching observations across parallel episodes. Each random proposal has the same nominal pre-projection radius as the corresponding \method{} update, although box projection shortens its realized displacement. The signed changes retain independently evaluated matched nominal-radius control estimates.

\emph{Metric and comparison conventions.} Table~\ref{tab:transfer-controls}(b) reports the canonical Base and signed effect estimates. For \method{}, Base${}+\Delta_R$ exactly reproduces Table~\ref{tab:main}. Bold marks the largest displayed effect within a row. The complete method and controls use different evaluation batches, making the ordering descriptive; MPE normalized points and SMAC shaped-return magnitudes remain within their respective benchmarks.

\noindent\emph{Result analysis.} The canonical \method{} gain is positive on $20/24$ settings and zero on \texttt{3m-Good}. Random directions are positive on several rows, including \texttt{Spread-Random} and \texttt{Tag-Expert}.

\noindent\emph{Conclusion.} The table compares the critic-guided effect with two matched nominal-radius perturbations. Separate batches yield descriptive control effects; across the two benchmarks, random movement and mismatched conditioning produce a different ordering from the complete method.

\FloatBarrier
\subsubsection{Qualitative Benchmark Context}

\paragraph{What behaviors and visual structures underlie the two benchmark families?}
This figure provides qualitative context for the action spaces and coordination patterns evaluated above. Quantitative performance evidence for Q1--Q3 appears in the corresponding tables and figures.

\emph{Selection protocol and figure conventions.} All panels use the highest-quality offline split. Episodes are ranked jointly by return and task-specific motion or completion statistics, and the fifth-ranked episode is displayed to avoid choosing only the single most favorable trajectory. Panels (a)--(c) show Spread-Expert (episode 327), Tag-Expert (episode 106), and World-Expert (episode 167). Panels (d)--(g) show 3m-Good (episode 834), 2s3z-Good (episode 732), 5m\_vs\_6m-Good (episode 32), and 8m-Good (episode 34). Allies are violet hexagons, and enemies are amber diamonds. Each row is one trajectory, and time advances from $0\%$ to $100\%$ from left to right. The rows display temporal motion; return differences are reported in the quantitative results.

\noindent\emph{Interpretation.} The rows expose the coordination structures referenced by the quantitative experiments: spatial coverage in Spread, joint pursuit in Tag and World, and focus fire or mixed-unit control in SMAC. They provide benchmark context across the different environments and selection procedures.

\begin{figure}[H]
\centering
\includegraphics[width=\textwidth]{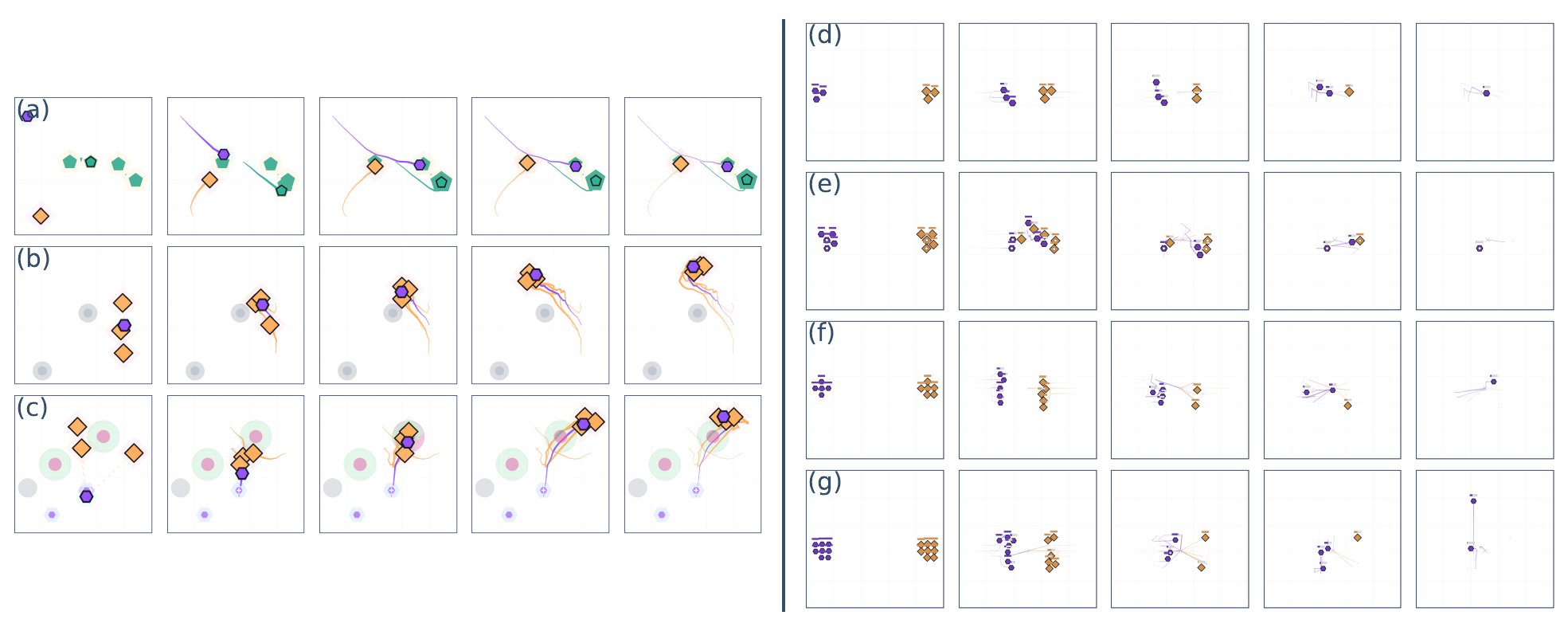}
\caption{\textbf{Qualitative benchmark trajectories.} Each row shows one trajectory at $0\%$, $25\%$, $50\%$, $75\%$, and $100\%$ progress. Panels (a)--(c) cover MPE and panels (d)--(g) cover SMAC.}
\label{fig:dataset-keyframes}
\end{figure}

\noindent\emph{Qualitative takeaway.} The trajectories make the coordination structures concrete: spatial coverage in Spread, joint pursuit in Tag and World, and focus fire or mixed-unit control in SMAC. They provide benchmark context rather than a performance comparison.

% \clearpage
\section{ Appendix B:  Algorithm and Mathematical Derivations}
\label{app:theory}

\subsection{Formal Multi-Agent Setting and Coordination Gap}
\label{app:formal-setting}

We formalize the task as a cooperative decentralized partially observable Markov decision process (Dec-POMDP)~\citep{oliehoek2016concise}:
\begin{equation}
\begin{aligned}
\mathcal M
&=\big(\mathcal N,\mathcal S,
\{\mathcal O_i\}_{i=1}^{n},
\{\mathcal A_i\}_{i=1}^{n},
P,\Omega,R,\gamma\big),\\
\mathcal O&=\prod_{i=1}^{n}\mathcal O_i,
\qquad \bm o_t=(\bm o_{t,i})_{i=1}^{n},\\
\mathcal A&=\prod_{i=1}^{n}\mathcal A_i,
\qquad \bm a_t=(\bm a_{t,i})_{i=1}^{n}.
\end{aligned}
\label{eq:ma_process}
\end{equation}
Here $\mathcal N=\{1,\ldots,n\}$ is the agent set, $x_t\in\mathcal S$ is the latent global environment state, $\bm o_{t,i}\in\mathcal O_i$ is agent $i$'s local observation generated through $\Omega$, and $\bm a_{t,i}\in\mathcal A_i$ is its local action. The joint action drives $P(x_{t+1}\mid x_t,\bm a_t)$ and receives the bounded shared reward $r_t=R(x_t,\bm a_t)$. The cooperative objective is $J(\pi)=\mathbb E_\pi[\sum_{t\ge0}\gamma^t r_t]$ with $\gamma\in[0,1)$. For continuous control, $\mathcal A_i\subset\mathbb R^{d_{a,i}}$ and $d_a=\sum_i d_{a,i}$; for SMAC, each $\mathcal A_i$ is a finite legal-action set. The policy and critic use $\bm o_t$, not the privileged latent state $x_t$.

The dataset is collected by an unknown joint behavior policy $\mu(\bm a\mid\bm o)$. A frozen generative joint policy $\pi_\theta:\mathcal O\times\mathcal G\to\Delta(\mathcal A)$ is trained on $\mathcal D$ and need not factorize over agents. The experiments use centralized test-time coordination execution: a coordinator evaluates $Q_\phi:\mathcal O\times\mathcal A\to\mathbb R$, refines every action block jointly, and dispatches $\tilde{\bm a}_i$ to agent $i$. This is centralized refinement followed by simultaneous multi-agent execution, not decentralized execution with the critic removed. No additional environment interaction or policy update is permitted.

Let $Q^\mu(\bm{o},\bm{a})$ denote the observation-conditioned expected discounted team return after taking $\bm a$ and then following $\mu$, and let $\mathcal{A}_{\mathcal D}(\bm{o})=\operatorname{supp}\mu(\cdot\mid\bm{o})$. Under the idealized support-matching assumption $\bm a_\theta\in\mathcal A_{\mathcal D}(\bm o)$ almost surely, define $\mathcal N_\theta(\bm o;r_{\rm loc})=\mathcal A_{\mathcal D}(\bm o)\cap\{\bm a:\|\bm a-\bm a_\theta\|_2\le r_{\rm loc}\}$ for $r_{\rm loc}>0$. The target-conditioned local coordination gap is
\begin{equation}
\begin{aligned}
\mathcal{G}_{r_{\rm loc}}(\bm{o},g)
&=\mathbb{E}_{\bm{a}_\theta}\!\left[
\sup_{\bm a\in\mathcal N_\theta(\bm o;r_{\rm loc})}
Q^\mu(\bm o,\bm a)\right.\\
&\hspace{6.5em}\left.{}-Q^\mu(\bm o,\bm a_\theta)
\right],
\qquad
\bm a_\theta\sim\pi_\theta(\cdot\mid\bm o,g).
\end{aligned}
\label{eq:gap}
\end{equation}
The support-matching assumption makes $\mathcal N_\theta$ nonempty and ensures $\mathcal G_{r_{\rm loc}}(\bm o,g)\ge0$. Whenever $\tilde{\bm a}\in\mathcal N_\theta$ and $Q_\phi=Q^\mu$ locally, increasing $Q_\phi(\bm o,\tilde{\bm a})$ reduces the samplewise gap by the same amount. Equation~\ref{eq:ccr} bounds the refinement displacement by $\eta$, and Eq.~\ref{eq:local_ascent_bound} establishes the corresponding critic-value increase. The experimental generator is specified in the later subsection \emph{The Coordinated Generative Backbone}.

\subsection{Centralized Behavior Critic Training}
\label{app:critic-training}
The centralized critic uses one-step fitted-Q evaluation of the empirical behavior policy:
\begin{equation}
\begin{aligned}
\mathcal L_Q(\phi)
&=\mathbb{E}_{\mathcal D_{\rm seq}}
\left[\big(Q_\phi(\bm o,\bm a)-y\big)^2\right],\\
y&=r+\gamma c\,\bar Q(\bm o',\bm a'),
\qquad
\phi^\star\in\arg\min_\phi\mathcal L_Q(\phi).
\end{aligned}
\label{eq:fqe}
\end{equation}
The data loader discards the final tuple of each episode because it has no logged successor action. Every retained pair therefore uses $c=1$ and bootstraps from its logged successor. The resulting critic is a behavior-continuation surrogate on retained data-supported pairs, not an exact finite-horizon return estimator at omitted episode endpoints.

\subsection{The \method{}  Algorithm}
\label{app:algorithm}
\method{} requires only a frozen joint policy and a differentiable centralized critic. The critic $Q_\phi(\bm{o},\bm{a})$ is fit once, offline, by Eq.~\ref{eq:fqe}; it involves no actor optimization.  At deployment both models are frozen, and each decision applies one globally normalized joint gradient ( Algorithm~\ref{alg:reco}). In continuous control this gradient is taken with respect to the executable action; in discrete control it is taken with respect to the concatenated per-agent logits through a masked softmax relaxation. In either case the normalization is over the \emph{whole joint vector}, rather than separately per agent, and bounds the pre-projection action or logit displacement by $\eta$.

\begin{algorithm}[H]
\caption{\method{}: test-time joint-action refinement (one decision)}
\label{alg:reco}
\begin{algorithmic}[1]
\REQUIRE frozen joint policy $\pi_\theta$; centralized critic $Q_\phi$; step size $\eta$; stabilizer $\varepsilon$; target return $g$
\STATE observe joint observation $\bm{o}$
\IF{continuous actions}
    \STATE $\bm a_\theta\sim\pi_\theta(\cdot\mid\bm o,g)$
    \STATE $\bm d\gets\nabla_{\bm a}Q_\phi(\bm o,\bm a_\theta)$
    \STATE $\tilde{\bm a}\gets\Pi_{\mathcal A}\!\left(\bm a_\theta+\eta\bm d/(\|\bm d\|_2+\varepsilon)\right)$
\ELSE
    \STATE obtain per-agent logits $\bm\ell_i$ from $\pi_\theta$ and legal masks $\bm m_i$ from the environment
    \STATE $\bm p_i\gets\mathrm{softmax}(\bm\ell_i+\log\bm m_i)$ for every agent; concatenate $\bm p$
    \STATE $\bm d_\ell\gets\nabla_{\bm\ell}Q_\phi(\bm o,\bm p)$
    \STATE $\tilde{\bm\ell}\gets\bm\ell+\eta\bm d_\ell/(\|\bm d_\ell\|_2+\varepsilon)$
    \STATE $\tilde a_i\gets\arg\max_j\{\tilde\ell_{ij}+\log m_{ij}\}$ for every agent
    \STATE $\tilde{\bm a}\gets(\tilde a_1,\ldots,\tilde a_n)$
\ENDIF
\STATE execute $\tilde{\bm a}$ \hfill\COMMENT{one critic backward pass; no rollout}
\end{algorithmic}
\end{algorithm}

\subsection{Discrete Joint  Actions}
\label{app:discrete-derivation}
Equation~\ref{eq:discrete} gives the complete masked-logit update. The critic is trained on concatenated one-hot actions, while $\bm p$ is the differentiable relaxation used to obtain the joint-logit gradient. The executable action is then re-masked and discretized, which preserves action legality.

\subsection{Derivation: \method{} as a Local Centralized Value Tilt}
\label{app:local-tilt}
Equations~\ref{eq:marl_kl_objective} and~\ref{eq:multi_qgf_policy} define the KL objective and its value-tilted optimizer. They give an exact \emph{statewise proximal improvement}. Assume that the frozen joint policy has a density (or a probability mass function in the discrete case), that the candidate $\pi$ is absolutely continuous with respect to $\pi_\theta$, that the terms in Eq.~\ref{eq:marl_kl_objective} are well defined, and that the partition function is finite. (For discrete actions the integral in Eq.~\ref{eq:multi_qgf_policy} is a sum.) Substitution gives the exact identity
\begin{equation}
\mathcal J_{\bm o,g}(\pi)
=\beta\log Z_\beta(\bm o,g)
-\beta D_{\rm KL}\!\left(
\pi(\cdot\mid\bm o,g)\,\|\,\pi_Q(\cdot\mid\bm o,g)
\right),
\label{eq:kl_completion}
\end{equation}
Nonnegativity of KL then proves that Eq.~\ref{eq:multi_qgf_policy} is the unique optimizer up to null sets. Equation~\ref{eq:multi_qgf_score} gives its continuous-density score. The centralized tilt is generally nonfactorized whenever the prior or $Q_\phi$ contains cross-agent interactions; even a factorized prior becomes coupled unless the value decomposes compatibly across agents. For the derivation, write $\bm d_\theta=\nabla_{\bm a}Q_\phi(\bm o,\bm a_\theta)=[\bm d_{\theta,1}^\top,\ldots,\bm d_{\theta,n}^\top]^\top$, where $\bm d_{\theta,i}=\nabla_{\bm a_i}Q_\phi(\bm o,\bm a_\theta)$. For the CGB, $\pi_\theta$ is an implicit push-forward. Drawing $\bm a_\theta\sim\pi_\theta$ supplies a behavior-supported initialization, and the following trust-region solution converts the Gibbs value tilt into a directly implementable first-order update:
\begin{equation}
\bm s^\star
=\arg\max_{\|\bm s\|_2\le\eta}
\langle\bm d_\theta,\bm s\rangle
=\eta\frac{\bm d_\theta}{\|\bm d_\theta\|_2},
\qquad \bm d_\theta\ne\bm0.
\label{eq:ccr_trust_solution}
\end{equation}

\paragraph{Interpretation of the joint normalized step.}
 At deployment, \method{} edits the numerical action vector that is about to be executed; it neither resamples a new action nor updates the parameters of $\pi_\theta$ or $Q_\phi$.  The critic derivative $\bm d_\theta=\nabla_{\bm a}Q_\phi(\bm o,\bm a_\theta)$ specifies the first-order direction in action space that raises the critic value most rapidly near the frozen proposal.  The constraint $\|\bm s\|_2\leq\eta$ assigns a fixed radius to this local edit.  Equation~\ref{eq:ccr_trust_solution} therefore selects the value-ascent direction while making $\eta$ directly control the largest permitted displacement of the complete joint action.

 All agent blocks are concatenated before computing $\|\bm d_\theta\|_2$.  The common denominator consequently preserves the relative magnitudes that the centralized critic assigns to different agents' corrections, while giving the team one shared correction budget.  Normalizing each $\bm d_{\theta,i}$ separately would instead give every agent its own radius-$\eta$ move; the resulting joint displacement could exceed $\eta$ and would discard the critic's relative allocation across agent blocks.  The stabilizer $\varepsilon$ makes the implemented denominator well defined when the gradient is zero or very small.  For the box-constrained continuous action sets used here, $\Pi_{\mathcal A}$ is coordinatewise Euclidean projection to the legal action bounds.  In particular, for $\mathcal A=\prod_j[a_j^{\min},a_j^{\max}]$,
\begin{equation}
[\Pi_{\mathcal A}(\bm x)]_j
=\min\!\left\{a_j^{\max},\max\!\left\{a_j^{\min},x_j\right\}\right\}.
\label{eq:box_projection}
\end{equation}
Thus the final vector remains executable even when the unconstrained step crosses a boundary.

For $\bm d_\theta=\bm0$ we set $\bm s^\star=\bm0$. Equation~\ref{eq:ccr} is the numerically stable, action-feasible version of Eq.~\ref{eq:ccr_trust_solution}: it replaces $\|\bm d_\theta\|_2$ by $D_\varepsilon=\|\bm d_\theta\|_2+\varepsilon$ and projects onto $\mathcal A$. Its agent blocks are
\begin{equation}
\tilde{\bm a}_i
=\Pi_{\mathcal A_i}\!\left(
\bm a_{\theta,i}+\eta\frac{\bm d_{\theta,i}}{D_\varepsilon}
\right),
\qquad i=1,\ldots,n .
\label{eq:ccr_blocks}
\end{equation}
Since projection onto a closed convex set is nonexpansive and $\bm a_\theta\in\mathcal A$,
\begin{equation}
\|\tilde{\bm a}-\bm a_\theta\|_2
\le
\eta\frac{\|\bm d_\theta\|_2}{\|\bm d_\theta\|_2+\varepsilon}
\le\eta.
\label{eq:ccr_trust}
\end{equation}
Let $\bm s=\tilde{\bm a}-\bm a_\theta$ and $\alpha=\eta/D_\varepsilon$. The variational inequality for Euclidean projection gives
\begin{equation}
\langle \bm d_\theta,\bm s\rangle
\ge \frac{1}{\alpha}\|\bm s\|_2^2
=\frac{D_\varepsilon}{\eta}\|\bm s\|_2^2.
\label{eq:projection_ascent}
\end{equation}
If $Q_\phi(\bm o,\cdot)$ has an $L$-Lipschitz gradient along the segment from $\bm a_\theta$ to $\tilde{\bm a}$, the smoothness inequality yields
\begin{equation}
\begin{aligned}
Q_\phi(\bm o,\tilde{\bm a})-Q_\phi(\bm o,\bm a_\theta)
&\ge \langle\bm d_\theta,\bm s\rangle-\frac{L}{2}\|\bm s\|_2^2\\
&\ge \left(\frac{D_\varepsilon}{\eta}-\frac{L}{2}\right)\|\bm s\|_2^2,
\end{aligned}
\label{eq:local_ascent_bound}
\end{equation}
Hence $0<\eta<2D_\varepsilon/L$ guarantees strict centralized critic-value improvement for every nonzero projected update. Under the local critic-consistency and support-preservation conditions stated after Eq.~\ref{eq:gap}, the same increase closes the samplewise coordination gap by an equal amount.

\subsection{The Coordinated Generative Backbone}
\label{app:cgb}
For experiments, $\pi_\theta$ is instantiated by a coordinated generative backbone (CGB) \citep{zou2026coflow}. This backbone is useful because it is a strong frozen generative policy and exposes a nontrivial value-injection question. The trajectory-space \method{} variants modify its internal trajectory, whereas the canonical action-space variant modifies its final emitted joint action. Write $\vartheta$ for the trajectory-flow parameters and $\psi$ for the inverse-dynamics parameters; the frozen policy shorthand is $\theta=(\vartheta,\psi)$. Let $\mathsf N_o$ be the fixed dataset observation normalizer; for continuous control, let $\mathsf N_a$ be the action normalizer and $\mathsf U_a=\mathsf N_a^{-1}$ denote action unnormalization. CGB models the clean, normalized joint observation trajectory $\bm{x}_0=\mathsf N_o(\bm{o}_t,\ldots,\bm{o}_{t+H})$ by transporting Gaussian noise $\bm{z}_1\!\sim\!\mathcal{N}(0,I)$ to $\bm{x}_0$ along the linear interpolant
\begin{equation}
\bm{z}_\alpha=(1-\alpha)\bm{x}_0+\alpha\bm{z}_1,\qquad \alpha\in[0,1],
\label{eq:coflow_interpolant}
\end{equation}
The base averaged-velocity regression term is Eq.~\ref{eq:coflow_loss}, with $\alpha\sim\mathcal U[0,1]$.
The released CGB checkpoints additionally use CoFlow's finite-difference consistency regularizer during training. The frozen sampling map and the \method{} formulas below use the resulting checkpoint directly. Given a schedule $1=\alpha_K>\alpha_{K-1}>\cdots>\alpha_0=0$, sampling runs from noise toward the data manifold with a few reverse Euler steps,
\begin{equation}
\begin{aligned}
\bm{z}^{k-1}
&=\bm{z}^{k}-(\alpha_k-\alpha_{k-1})\,
u_\vartheta(\bm{z}^{k},0,\alpha_k;\mathsf N_o(\bm{o}),g),\\
\bm{z}^{K}&\sim\mathcal{N}(0,I),
\end{aligned}
\label{eq:coflow_sampling}
\end{equation}
or by the one-step shortcut $\hat{\bm{x}}_0=\bm{z}^K-u_\vartheta(\bm{z}^K,0,1;\mathsf N_o(\bm{o}),g)$. The velocity is natively joint. The backbone decomposes each agent's velocity into an individual term and a coordination term,
\begin{equation}
u_\vartheta^i=u_{\mathrm{ind}}^i+u_{\mathrm{coord}}^i ,
\label{eq:coflow_cva}
\end{equation}
and computes the coordination term by Coordinated Velocity  Attention (CVA ).  At layer $l$, let $\bm{c}_l^i$ be agent $i$'s temporal feature. Shared projections form query, key, and value features,
\begin{equation}
\bm{q}_l^i=W_Q\bm{c}_l^i,\qquad
\bm{\kappa}_l^j=W_K\bm{c}_l^j,\qquad
\bm{v}_l^j=W_V\bm{c}_l^j .
\label{eq:cva_qkv}
\end{equation}
 Agent $i$ attends to all agents $j$ through
\begin{equation}
\begin{aligned}
\omega_{ij}^{(l)}
&=
\frac{\exp((\bm{q}_l^i)^\top \bm{\kappa}_l^j/\sqrt{d_{\rm att}})}
{\sum_{m=1}^{n}\exp((\bm{q}_l^i)^\top \bm{\kappa}_l^m/\sqrt{d_{\rm att}})},\\
\bm{\xi}_l^i&=\sum_{j=1}^{n}\omega_{ij}^{(l)}\bm{v}_l^j,
\end{aligned}
\label{eq:cva_attention}
\end{equation}
and injects the teammate message through a learnable coordination gate,
\begin{equation}
\hat{\bm{c}}_l^i=\bm{c}_l^i+\chi_l\bm{\xi}_l^i ,
\label{eq:cva_gate}
\end{equation}
where $\bm\kappa_l^j$ is a key vector, $d_{\rm att}$ is its dimension, $\bm\xi_l^i$ is the aggregated teammate message, and $\chi_l$ is the learnable coordination gate. The symbols $\bm\kappa_l^j$ and $\bm\xi_l^i$ avoid overloading the denoising-step index $k$ and the legal-action mask $\bm m_i$, respectively. The value projection has output dimension matching $\bm c_l^i$, making the residual sum in Eq.~\ref{eq:cva_gate} dimensionally valid. The shared attention weights let each agent's velocity depend on teammate features at the same denoising layer; $\chi_l$ controls the strength of this cross-agent message and is distinct from the discount $\gamma$. Thus the generated trajectory is coupled before inverse dynamics is applied. Let $I_\psi$ be the shared per-agent inverse-dynamics head and $\mathcal I_\psi$ its componentwise joint lifting. For continuous control, the head outputs a normalized joint action, which is unnormalized before execution:
\begin{equation}
\begin{aligned}
\bm{a}^{\rm norm}_\theta
&=\mathcal I_\psi(\hat{\bm o}^{\rm norm}_t,
\hat{\bm o}^{\rm norm}_{t+1})
:=\big(
I_\psi(\hat{\bm o}^{1,\rm norm}_t,
\hat{\bm o}^{1,\rm norm}_{t+1}),\ldots,
I_\psi(\hat{\bm o}^{n,\rm norm}_t,
\hat{\bm o}^{n,\rm norm}_{t+1})
\big),\\
\bm a_\theta&=\mathsf U_a(\bm a^{\rm norm}_\theta).
\end{aligned}
\label{eq:inverse_dynamics}
\end{equation}
For discrete control the same head emits logits and Eq.~\ref{eq:discrete} supplies masking and discretization in place of $\mathsf U_a$. The frozen policy used in the experiments is the push-forward of Gaussian noise through a coordinated trajectory generator, inverse-dynamics decoder, and output transform. Writing $F_\vartheta(\bm{z}^K;\mathsf N_o(\bm{o}),g)=\hat{\bm{x}}_0$ for the few-step sampler and $P_0,P_1$ for the first two generated normalized-observation slices, the continuous-action map is
\begin{equation}
\begin{aligned}
h_{\vartheta,\psi}(\bm{z}^K;\bm{o},g)
&=\mathsf U_a\!\left(\mathcal I_\psi\!\left(
P_0F_\vartheta(\bm{z}^K;\mathsf N_o(\bm{o}),g),
P_1F_\vartheta(\bm{z}^K;\mathsf N_o(\bm{o}),g)\right)\right),\\
\bm{z}^K&\sim\mathcal{N}(0,I),\qquad
\bm{a}_\theta=h_{\vartheta,\psi}(\bm{z}^K;\bm{o},g).
\end{aligned}
\label{eq:coflow_pushforward}
\end{equation}
The induced action policy is the conditional push-forward $\pi_\theta(\cdot\mid\bm{o},g)=\big(h_{\vartheta,\psi}(\cdot;\bm o,g)\big)_\#\mathcal{N}(0,I)$ with $\theta=(\vartheta,\psi)$. Cross-agent velocity attention allows each decoded action to depend on teammate features. The induced prior therefore need not be a product of conditionally independent per-agent policies:
\begin{equation}
\pi_\theta(\bm{a}\mid\bm{o},g)
\ \text{need not factorize as}\ 
\prod_{i=1}^{n}\pi_{\theta,i}(\bm a_i\mid\bm o,g).
\label{eq:nonfactorized_prior}
\end{equation}
Two properties matter for \method{}: (i) the executed action is the \emph{decoded} quantity $\bm{a}_\theta$, where the behavior critic's value signal is applied most directly; and (ii) the flow is \emph{short}. Perturbing intermediate iterates $\bm{z}^k$ is therefore a stronger intervention than in many-step diffusion policies and must be treated as a separate design choice (Sec.~Method).

\subsection{\method{} Value-Injection Variants for Coordinated Generators}
\label{app:injection-variants}
\method{} uses the same frozen coordinated policy and test-time critic guidance at three injection locations. All-step and final-step trajectory-space \method{} modify an internal denoising trajectory before action decoding; post-generation action-space \method{} modifies the decoded action after the flow is complete. For a generic action-space flow $v_\zeta$ under the present convention (data at $\alpha=0$, noise at $\alpha=1$), a one-step clean estimate and its action-space guidance direction are
\begin{equation}
\begin{aligned}
\hat{\bm a}_0(\alpha)
&=\bm a_\alpha-\alpha v_\zeta(\bm a_\alpha,\alpha;\bm o,g),\\
\bm G_{\rm act}(\alpha)
&=\nabla_{\hat{\bm a}_0}Q_\phi(\bm o,\hat{\bm a}_0),
\qquad
\widehat{\bm G}_{\rm act}
=\frac{\bm G_{\rm act}}{\|\bm G_{\rm act}\|_2+\varepsilon},\\
\bm a_{\alpha-\Delta\alpha}
&=\bm a_\alpha-\Delta\alpha\left[
v_\zeta(\bm a_\alpha,\alpha;\bm o,g)
-\lambda_Q\widehat{\bm G}_{\rm act}
\right],
\end{aligned}
\label{eq:inloop_qgf}
\end{equation}
where $\lambda_Q>0$ and $0<\Delta\alpha\le\alpha$. This reference update normalizes the guidance direction and uses $\lambda_Q$ as its displacement in denoising coordinates. Direct use of $\bm G_{\rm act}$ assumes $\partial\hat{\bm a}_0/\partial\bm a_\alpha\approx\operatorname{Id}$; the exact gradient premultiplies it by that Jacobian transpose. The outer minus sign yields value ascent under the reverse-time Euler convention.

For CGB, $\bm z^k$ is a normalized joint-observation trajectory, not an action. An identity map between these spaces is therefore invalid. The trajectory-space \method{} update takes the ordinary reverse-Euler step in Eq.~\ref{eq:coflow_intermediate_action}, then differentiates the normalized critic through the trajectory-to-observation--action decoder by Eq.~\ref{eq:coflow_chain_guidance}. The critic $Q_\nu^{\rm norm}$ uses the FQE loss in Eq.~\ref{eq:fqe} on identically normalized logged transitions. The implementation detaches $\bar{\bm z}^{k-1}$; this gradient includes the decoder and inverse-dynamics paths but not earlier denoising steps. The all-step and final-step trajectory-space \method{} variants use
\begin{equation}
\begin{aligned}
\widehat{\bm G}_{\rm traj}^{k-1}
&=\frac{\bm G_{\rm traj}^{k-1}}{\|\bm G_{\rm traj}^{k-1}\|_2+\varepsilon},\\
\bm z^{k-1}
&=\bar{\bm z}^{k-1}+\lambda_Q\widehat{\bm G}_{\rm traj}^{k-1}.
\end{aligned}
\label{eq:qgf_variants_coflow}
\end{equation}
The all-step variant applies Eq.~\ref{eq:qgf_variants_coflow} after every base Euler step; the final-step variant applies it only after the last one. Here $\lambda_Q$ is a normalized trajectory-space displacement with its own scale, independent of $\Delta\alpha_k$.

\paragraph{The decoder separates the \method{} variants.}
For Final, $k=1$: the last base step produces $\bar{\bm z}^{0}$, the normalized critic changes it to $\bm z^{0}$ through Eq.~\ref{eq:qgf_variants_coflow}, and the decoder maps $\bm z^{0}$ to $\tilde{\bm a}$.  Its gradient follows $Q_\nu^{\rm norm}\rightarrow\bm y^0\rightarrow\bar{\bm z}^{0}$ through the exact decoder Jacobian; All repeats this path at every step. Post completes the flow, decodes $\bm a_\theta$, and applies Eq.~\ref{eq:ccr} in executable action space after generation ends. In general $h_{\rm dec}(\bm z^0+\Delta\bm z)\neq h_{\rm dec}(\bm z^0)+\Delta\bm a$: a nonlinear decoder can rotate or rescale the trajectory-space direction. This coordinate effect reflects the decoder's coordinates. The experiment measures the resulting return differences across the two locations.

\end{document}